%% file: main.tex
\documentclass{article}
\PassOptionsToPackage{numbers}{natbib}
\usepackage[preprint]{neurips_2025}

\usepackage{amsmath}
\usepackage{bm}
\usepackage{xspace}
\usepackage{xcolor}
\usepackage{colortbl}
\usepackage{pifont}
\usepackage{bm}
\usepackage{rotating}
\usepackage{booktabs}
\usepackage{tikz}
\usetikzlibrary{positioning}
\usepackage{multicol}
\usepackage{multirow,array,xcolor,booktabs}
\usepackage[utf8]{inputenc} 
\usepackage[T1]{fontenc}    
\usepackage{hyperref}       
\usepackage{url}            
\usepackage{booktabs}       
\usepackage{amsfonts}       
\usepackage{nicefrac}       
\usepackage{microtype}      
\usepackage{xcolor}         
\usepackage{placeins}
\usepackage[hypcap=false]{caption}
\usepackage{floatrow}
\DeclareFloatSeparators{mysep}{\hskip-3em}

\input{preamble}

\input{sec/tables}
\input{sec/figures}

\title{\epsSeg: Sparsely Supervised Semantic Segmentation\\of Microscopy Data}

\author{%
  Sheida Rahnamai Kordasiabi\textsuperscript{1,2},\;
  Damian Dalle Nogare\textsuperscript{1},\;
  Florian Jug\textsuperscript{1} \\
  \textsuperscript{1}Human Technopole, Milan, Italy \\
  \textsuperscript{2}Technical University of Dresden, Germany
}

\begin{document}
\maketitle
\input{sec/0_abstract}    
\input{sec/1_intro}
\input{sec/2_related_work}
\input{sec/3_method}
\input{sec/4_results}
\input{sec/5_discussion}
\medskip
{
    \small
    \bibliographystyle{ieeenat_fullname}
    \bibliography{main}
}
\clearpage
\input{supplement}

\end{document}

%% file: preamble.tex
\newcommand{\miniheadline}[1]{\vspace{1.2mm}\noindent\textbf{#1.}}

\DeclareRobustCommand\onedot{\futurelet\@let@token\@onedot}
\def\onedot{. }
\def\eg{\emph{e.g}\onedot} 
\def\ie{\emph{i.e}\onedot}

\def\wrt{w.r.t\onedot}

\newcommand{\epsSeg}{\ensuremath{\bm\epsilon}-Seg\xspace}

\newcommand{\VAE}{\mbox{\textsc{VAE}}\xspace}
\newcommand{\VAEs}{\mbox{\textsc{VAE}}s\xspace}

\newcommand{\HVAE}{\mbox{\textsc{HVAE}}\xspace}
\newcommand{\HVAEs}{\mbox{\textsc{HVAEs}}\xspace}

\newcommand{\MLP}{\mbox{\textsc{MLP}}\xspace}

\newcommand{\GMM}{\mbox{\textsc{GMM}}\xspace}
\newcommand{\CL}{\mbox{\textsc{CL}}\xspace}
\newcommand{\EM}{\mbox{\textsc{EM}}\xspace}

\newcommand{\UNet}{\mbox{U-Net}\xspace}

\usepackage{amsmath}
\usepackage{graphicx}

%% file: sec/tables.tex
\newcommand\tabBaselineSupplement{
\begin{table*}
    \centering
    \resizebox{0.95\textwidth}{!}{ 
        \rowcolors{2}{gray!20}{}
        \begin{tabular}{lcccccc}
            \hline
            \textbf{Model} & \textbf{Learning Paradigm} & \textbf{U} & \textbf{N} & \textbf{G} & \textbf{M} & \textbf{Avg DSC} \\ 
            \hline
            Vanilla HVAE$^*$~\cite{hdn} & Self-Supervised & 0.44 & 0.55 & 0.34 & 0.13 & 0.37 \\ 
            Labkit~\cite{labkit} & Sparsely Supervised & 0.85 & 0.44 & 0.68 & 0.61 & 0.65 \\
            U-net~\cite{unet} & Fully Supervised & 0.94 & 0.99 & 0.90 & 0.87 & 0.93 \\
            U-net & Sparsely Supervised & 0.90 & 0.96 & 0.78 & 0.66 & 0.83 \\
            Vanilla ViT~\cite{vit} & Fully Supervised & 0.91 & 0.98 & 0.77 & 0.87 & 0.88 \\
            Segmenter~\cite{segmenter} & Fully Supervised & 0.91 & 0.99 & 0.86 & 0.90 & 0.92 \\
            MAESTER$^*$~\cite{maester} & Self-Supervised & 0.84 & 0.95 & 0.56 & 0.79 & 0.79 \\
            Han et al$^*$~\cite{han} & Self-Supervised & - & - & - & - & 0.66 \\
            \epsSeg (+$\mathcal{L}_H$) & Sparsely Supervised & 0.89 & 0.98 & 0.81 & 0.83 & 0.88  \\
            \hline
        \end{tabular}
    }
    \captionof{table}{Dice similarity coefficient per class and average across all classes comparing our model with baselines on the ``BetaSeg'' dataset~\cite{Muller2021-mg}. 
    Methods marked with an asterisk use K-Means clustering on latent features to conduct semantic segmentation (more explanation can be found in Section~\ref{mini:seg}). U: Unrecognized, N:Nucleus, G:Granules, M:Mitochondria.}
    \label{tab:baseline_sup}
\end{table*}
}

\newcommand\tabBaseline{
\begin{table*}
    \centering
    \begin{tabular}{l l c c c c c}
        \hline
        \textbf{Learning Paradigm} & \textbf{Model} & \textbf{U} & \textbf{N} & \textbf{G} & \textbf{M} & \textbf{Avg DSC} \\
        \hline
        \multirow{3}{*}{Self-Supervised}
        & Vanilla HVAE$^*$~\cite{hdn} & 0.44 & 0.55 & 0.34 & 0.13 & 0.37 \\
        & Han et al.$^*$~\cite{han}   &  --  &  --  &  --  &  --  & 0.66 \\
        & MAESTER$^*$~\cite{maester}  & 0.84 & 0.95 & 0.56 & 0.79 & 0.79 \\
        \hline
        \multirow{3}{*}{Sparsely Supervised}
        & Labkit~\cite{labkit}        & 0.85 & 0.44 & 0.68 & 0.61 & 0.65 \\
        & \UNet                       & 0.90 & 0.96 & 0.78 & 0.66 & 0.83 \\
        & \textbf{\epsSeg\ (ours)}    & 0.91 & 0.96 & 0.82 & 0.86 & 0.89 \\
        \hline
        \multirow{3}{*}{Fully Supervised}
        & Vanilla ViT~\cite{vit}      & 0.91 & 0.98 & 0.77 & 0.87 & 0.88 \\
        & Segmenter~\cite{segmenter}  & 0.91 & 0.99 & 0.86 & 0.90 & 0.92 \\
        & \UNet~\cite{unet}           & 0.94 & 0.99 & 0.90 & 0.87 & 0.93 \\
        \hline
    \end{tabular}
    \caption{Dice similarity coefficient per class and average across all classes on the ``BetaSeg'' dataset~\cite{Muller2021-mg}. 
    Methods marked with an asterisk use K-Means clustering on latent features to conduct semantic segmentation (see Section~\ref{mini:seg}). U: Unrecognized, N: Nucleus, G: Granules, M: Mitochondria.}
    \label{tab:baseline}
\end{table*}
}

\newcommand\tabLabel{
\begin{minipage}{0.45\textwidth}
    \centering
    \scriptsize
    \rowcolors{4}{gray!20}{}
    \begin{tabular}{lccccc}
        \hline
        \textbf{RLF} & \multicolumn{4}{c}{\textbf{Per-Class Dice Coefficient}} & \textbf{Avg} \\ 
        \cline{2-5}
        & \textbf{U} & \textbf{N} & \textbf{G} & \textbf{M} & \textbf{DSC} \\ 
        \hline
        20 & 0.89 & 0.98 & 0.81 & 0.83 & 0.88 \\ 
        15 & 0.88 & 0.98 & 0.81 & 0.78 & 0.86 \\ 
        10 & 0.86 & 0.98 & 0.80 & 0.75 & 0.85 \\ 
        5  & 0.85 & 0.96 & 0.77 & 0.76 & 0.84 \\ 
        1  & 0.79 & 0.95 & 0.69 & 0.69 & 0.78 \\ 
        \hline
    \end{tabular}
    \captionof{table}{DSC per class and average across all classes. The ``RLF'' column (Relative Labeling Factor) specify a scaling factor where 20 corresponds to 0.05\% and 1 as small as 0.0025\% of the total labeles available. U: Unrecognized, N:Nucleus, G:Granules, M:Mitochondria.}
    \label{tab:label}
\end{minipage}
}

\newcommand\tabLabelReb{
\begin{table}
    \centering
    \begin{tabular}{llccccc}
        \hline
        \textbf{RLF} & \textbf{Model} & \textbf{U} & \textbf{N} & \textbf{G} & \textbf{M} & \textbf{Avg DSC} \\
        \hline
        20 & U-net & 0.63 & 0.75 & 0.51 & 0.12 & 0.50 \\
           & \epsSeg & 0.89 & 0.98 & 0.81 & 0.83 & 0.88 \\
        \hline
        15 & U-net & 0.53 & 0.64 & 0.41 & 0.14 & 0.43 \\
           & \epsSeg & 0.88 & 0.98 & 0.81 & 0.78 & 0.86 \\
        \hline
        10 & U-net & 0.30 & 0.20 & 0.42 & 0.34 & 0.31 \\
           & \epsSeg & 0.86 & 0.98 & 0.80 & 0.75 & 0.85 \\
        \hline
         5 & U-net & 0.71 & 0.00 & 0.00 & 0.03 & 0.18 \\
           & \epsSeg & 0.85 & 0.96 & 0.77 & 0.76 & 0.84 \\
        \hline
         1 & U-net & 0.17 & 0.00 & 0.37 & 0.02 & 0.14 \\
           & \epsSeg & 0.79 & 0.95 & 0.69 & 0.69 & 0.78 \\
        \hline
    \end{tabular}
    \captionof{table}{Comparison between U-Net and \epsSeg on the ``BetaSeg'' dataset under varying label sparsity levels. 
    ``RLF'' (Relative Labeling Factor) specifies the fraction of available labels, where 20 corresponds to 0.05\% and 1 to 0.0025\% of total labels. 
    U: Unrecognized, N: Nucleus, G: Granules, M: Mitochondria. Although both models were trained with \textit{balanced supervision}, using patches selected to include all classes, 
    the U-Net still fails to segment the nucleus at very low labeling levels (RLF 1 and 5). 
    This illustrates a key limitation of discriminative models such as U-Net, under extreme supervision sparsity, 
    even balanced examples may not suffice to generalize fine-grained or context-sensitive structures like the nucleus. 
    In contrast, \epsSeg benefits from its class-aware latent modeling via the GMM prior, 
    which enables it to extract meaningful representations for different structures and distinguish them semantically.
    We note that the sparse U-Net reported earlier was trained on slice numbers 800, 600, and 500 
    of the “high\_c1”, “high\_c2”, and “high\_c3” volumes of the ``BetaSeg'' dataset. For selecting the same amount of data used in \epsSeg, to train the 2D U-Net on, as reported in the table above, we extracted 64x64 patches where except background, different classes are approximately well balanced.}
    \label{tab:unet_reb}
\end{table}
}

\newcommand\tabResBlock{
\begin{table}
    \centering
    \rowcolors{5}{gray!20}{}
    \begin{tabular}{lccccc}
        \hline
        \textbf{\# res.} & \multicolumn{4}{c}{\textbf{Per-Class Dice Coefficient}} & \textbf{Avg} \\
        \cline{2-5}
        \textbf{blocks} & \textbf{U} & \textbf{N} & \textbf{G} & \textbf{M} & \textbf{DSC} \\
        \hline
        5 & 0.86 & 0.98 & 0.80 & 0.75 & 0.85 \\
        4 & 0.85 & 0.97 & 0.80 & 0.74 & 0.84 \\
        3 & 0.88 & 0.96 & 0.81 & 0.80 & 0.86 \\
        2 & 0.87 & 0.97 & 0.81 & 0.77 & 0.86 \\
        1 & 0.85 & 0.97 & 0.80 & 0.72 & 0.84 \\
        \hline
    \end{tabular}
    \captionof{table}{Residual blocks ablation (3 latent variables). U: Unrecognized, N: Nucleus, G: Granules, M: Mitochondria.}
    \label{tab:resblock}
\end{table}
}

\newcommand\tabLatent{
\begin{table}
    \centering
    \begin{tabular}{lccccc}
        \hline
        \textbf{\# latent} & \multicolumn{4}{c}{\textbf{Per-Class Dice Coefficient}} & \textbf{Avg} \\ 
        \cline{2-5}
        & \textbf{U} & \textbf{N} & \textbf{G} & \textbf{M} & \textbf{DSC} \\ 
        \hline
        2          & 0.87 & 0.98 & 0.81 & 0.76 & \textbf{0.86} \\ 
        3          & 0.86 & 0.98 & 0.80 & 0.75 & 0.85 \\ 
        \hline
    \end{tabular}
    \captionof{table}{Latent variables ablation (5 res. blocks/layer). U: Unrecognized, N:Nucleus, G:Granules, M:Mitochondria.}
    \label{tab:latent}
\end{table}
}

\newcommand\tabEntropy{
\begin{minipage}{0.45\textwidth}
    \centering
    \scriptsize
    \begin{tabular}{lccccc}
        \hline
        \textbf{Entropy} & \multicolumn{4}{c}{\textbf{Per-Class Dice Coefficient}} & \textbf{Avg} \\
        \cline{2-5}
        \textbf{Loss} & \textbf{U} & \textbf{N} & \textbf{G} & \textbf{M} & \textbf{DSC} \\
        \hline
        \ding{55} & 0.81 & 0.97 & 0.74 & 0.71 & 0.81 \\
        \ding{51} & 0.86 & 0.98 & 0.80 & 0.75 & \textbf{0.85} \\
        \hline
    \end{tabular}
    \captionof{table}{Effect of entropy loss: The best checkpoint of a sparsely supervised model was further trained using batches with 50\% unlabeled data. U: Unrecognized, N:Nucleus, G:Granules, M:Mitochondria.}
    \label{tab:entropy}
\end{minipage}
}

\newcommand\tabLoss{
\begin{minipage}{0.45\textwidth}
    \resizebox{0.99\textwidth}{!}{ 
        \begin{tabular}{lcccccccc}
            \hline
            &\textbf{Loss} & &\textbf{Prior}&\multicolumn{4}{c}{\textbf{Per-Class Dice Coefficient}} & \textbf{Avg} \\ 
            \cline{1-3}
            \cline{5-8}
            \textbf{KL} & \textbf{CL} & \textbf{CE} & \textbf{Distribution} & \textbf{U} & \textbf{N} & \textbf{G} & \textbf{M} & \textbf{DSC} \\ 
            \hline
             \ding{51} & \ding{55} & \ding{55} & $\mathcal{N}$ & 0.44 & 0.55 & 0.34 & 0.13 & 0.37 \\
             \rowcolor{gray!20}
             \ding{51} & \ding{51} & \ding{55} & $\mathcal{N}$ & 0.83 & 0.95 & 0.69 & 0.76 & 0.81 \\ 
             \ding{51} & \ding{55} & \ding{51} & $\mathcal{N}$ & 0.81 & 0.97 & 0.80 & 0.75 & 0.83 \\
             \rowcolor{gray!20}
             \ding{51} & \ding{51} & \ding{51} & $\mathcal{N}$ & 0.81 & 0.97 & 0.73 & 0.72 & 0.81 \\ 
             \ding{51} & \ding{55} & \ding{51} & GMM & 0.82 & 0.97 & 0.72 & 0.75 & 0.82 \\
             \ding{51} & \ding{51} & \ding{51} & GMM & 0.86 & 0.98 & 0.80 & 0.75 & \textbf{0.85} \\ 
            \hline
        \end{tabular}
    }
    \captionof{table}{Loss components and prior distribution ablation on ``BetaSeg'' dataset. U: Unrecognized, N: Nucleus, G: Granules, M: Mitochondria.}
    \label{tab:loss}
\end{minipage}
}

\newcommand\tabMasking{
\begin{minipage}{0.45\textwidth}
    \resizebox{0.99\textwidth}{!}{ 
        \centering
        \rowcolors{4}{gray!20}{}
        \begin{tabular}{lccccc}
            \hline
            \textbf{Mask} & \multicolumn{4}{c}{\textbf{Per-Class Dice Coefficient}} & \textbf{Avg} \\ 
            \cline{2-5}
            \textbf{Size} & \textbf{Unrecognized} & \textbf{Nucleus} & \textbf{Granules} & \textbf{Mitochondria} & \textbf{DSC} \\ 
            \hline
            9x9          & 0.83 & 0.95 & 0.65 & 0.73 & 0.79 \\ 
            7x7          & 0.84 & 0.97 & 0.72 & 0.75 & 0.82 \\ 
            5x5          & 0.87 & 0.94 & 0.78 & 0.80 & 0.85 \\ 
            3x3          & 0.88 & 0.97 & 0.81 & 0.80 & \textbf{0.87} \\ 
            1x1          & 0.86 & 0.98 & 0.80 & 0.75 & 0.85 \\ 
            \hline
        \end{tabular}
    }
    \captionof{table}{Label consistency ablation on ``BetaSeg'' dataset. The “Mask Size” column indicates the size of the center-region mask, within which the pixel-wise ground truth labels are consistent.}
        \label{tab:mask}
\end{minipage}
}

\newcommand\tabLabkit{
\begin{minipage}{0.45\textwidth}
    \centering
    \scriptsize
    \rowcolors{4}{gray!20}{}
    \begin{tabular}{lccccc}
        \hline
        \textbf{Trained} & \multicolumn{4}{c}{\textbf{Per-Class Dice Coefficient}} & \textbf{Avg} \\ 
        \cline{2-5}
        \textbf{on} & \textbf{U} & \textbf{N} & \textbf{G} & \textbf{M} & \textbf{DSC} \\ 
        \hline
        high\_c1 & \textbf{0.85} & 0.38 & \textbf{0.68} & \textbf{0.61} & 0.63 \\ 
        high\_c2 & 0.80 & 0.33 & 0.58 & 0.56 & 0.57 \\ 
        high\_c3 & 0.82 & \textbf{0.44} & 0.63 & 0.42 & 0.58 \\ 
        \hline
    \end{tabular}
    \captionof{table}{Labkit results. Due to different image sizes, Labkit was trained on individual volumes. U: Unrecognized, N: Nucleus, G: Granules, M: Mitochondria.}
    \label{tab:labkit}
\end{minipage}
}

\newcommand\tabFlu{
\begin{minipage}{0.45\textwidth}
    \centering
    \scriptsize
    \begin{tabular}{cccc}
        \hline
        \multicolumn{3}{c}{\textbf{Per-Class Dice Coefficient}} & \textbf{Avg} \\ 
        \cline{1-3}
        \textbf{Background} & \textbf{Cytoplasm} & \textbf{Nuclei} & \textbf{DSC} \\ 
        \hline
        0.94 & 0.86 & 0.90 & 0.90 \\ 
        \hline
    \end{tabular}
    \captionof{table}{Dice similarity coefficient per class and average for ``Aitslab-bioimaging'' datasets.}
    \label{tab:flu}
\end{minipage}
}

\newcommand\tabZerial{
\begin{minipage}{0.95\textwidth}
    \centering
    \resizebox{0.95\textwidth}{!}{ 
        \rowcolors{2}{gray!20}{}
        \begin{tabular}{lcccccccc}
            \hline
            \textbf{Model} & \textbf{B} & \textbf{M} & \textbf{P} & \textbf{L} & \textbf{BM} & \textbf{OBC} & \textbf{CBC} & \textbf{Avg DSC} \\ 
            \hline
            U-net~\cite{unet}-Fully Supervised & 0.97 & 0.95 & 0.85 & 0.79 & 0.52 & 0.87 & 0.90 & 0.84\\
            U-net-Sparsely Supervised          & 0.94 & 0.81 & 0.68 & 0.81 & 0.49 & 0.39 & 0.00 & 0.59 \\
            \epsSeg-Sparsely Supervised & 0.91 & 0.82 & 0.63 & 0.81 & 0.39 & 0.70 & 0.46 & 0.67  \\
            \hline
        \end{tabular}
    }
    \captionof{table}{Dice similarity coefficient per class and average across all classes comparing our model with baselines for ``liver FIBSEM'' dataset. B: Background, M: Mitochondria, P: Peroxisomes, L: Lipofuscin, BM: Basolateral Membrane, OBC: Open Bile Canaliculus, CBC: Closed Bile Canaliculus.}
    \label{tab:zerial}
\end{minipage}
}

\newcommand\tabComboLabelEntropy{
    \tabLabel \hspace{0.05\textwidth} \tabEntropy
}

\newcommand\tabComboFluLabkit{
    \tabFlu \hspace{0.05\textwidth} \tabLabkit 
}

\newcommand\tabComboLossMasking{
  \tabLoss \hspace{0.05\textwidth} \tabMasking
}

%% file: sec/figures.tex
\newcommand\figMain{
\begin{figure*}
    \centering
    \includegraphics[width=0.99\textwidth]{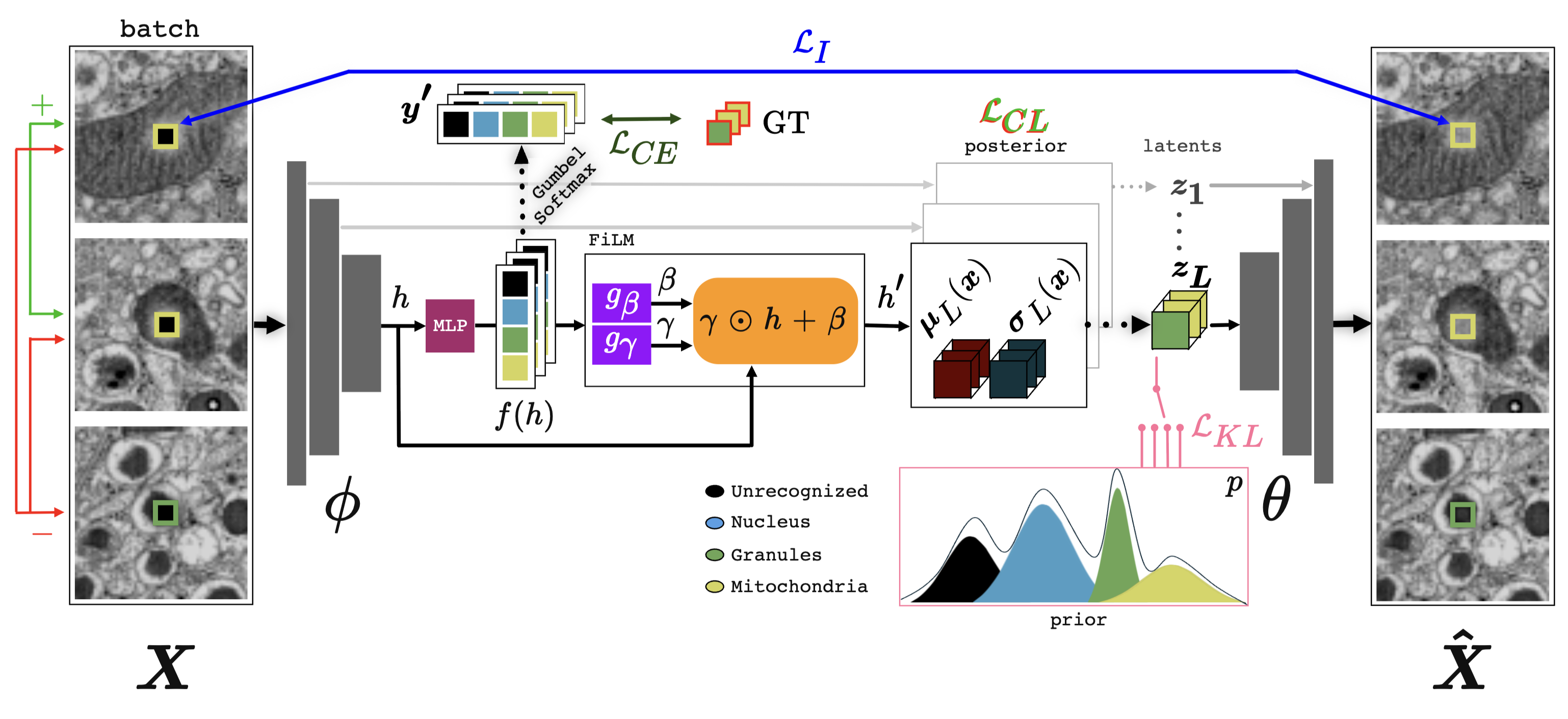}
    \caption{The overall pipeline of \epsSeg which is trained on an inpainting task (of center-region masked inputs). $\phi$ and $\theta$ are encoder and decoder of the network, respectively. Dotted arrows show sampling from a distribution (Gumbel-Softmax (Categorical-like distribution) for segmentation head and Normal distribution for conditional posterior). $h$ is an intermediate feature embedding of input $x$ coming from the encoder $\phi$. $f(h)$ is a logit vector and $|f(h)|=C$ with $C$ being the number of different classes/GMM prior components (equal to 4 for ``BetaSeg''~\cite{Muller2021-mg}). $\beta$ and $\gamma$ are feature-wise linear modulation (FiLM's~\cite{film}) parameters (shifting and scaling factors) of features $h$. $h'$ are the posterior distribution's parameters and are divided into two chunks shown as $\bm\mu_{\bm{L}}(\bm{x})$ and $\bm\sigma_{\bm{L}}(\bm{x})$ by c being the corresponding label of the masked center region of each input patch $x$ in the batch. $z_L$ is a sample from $\mathcal{N}(\bm\mu_{\bm{L}}(\bm{x}),\bm\sigma_{\bm{L}}^2(\bm{x}))$. $\bm{y}'$ is a differentiable sample from a Gumbel-Softmax~\cite{gumbel} distribution. Green arrow shows positive pair of patches having similar labels, and red arrows show negative pairs of patches having dissimilar labels. $\mathcal{L}_{CL}$ is then computed on $\bm\mu s$ (further explanation can be found in section ~\ref{mini:cl}). For $\mathcal{L}_{I}$ inpainting loss, $\mathcal{L}_{CL}$ contrastive loss, $\mathcal{L}_{CE}$ cross-entropy loss and $\mathcal{L}_{KL}$ refer to Equations~\ref{eq:mse_inpainting_loss},~\ref{eq:cl},~\ref{eq:ce} and~\ref{eq:kl} respectively.}
    \label{fig:main}
\end{figure*}
}
\newcommand\figBase{
\begin{figure}[hb]
    \centering
    \includegraphics[width=0.99\textwidth]{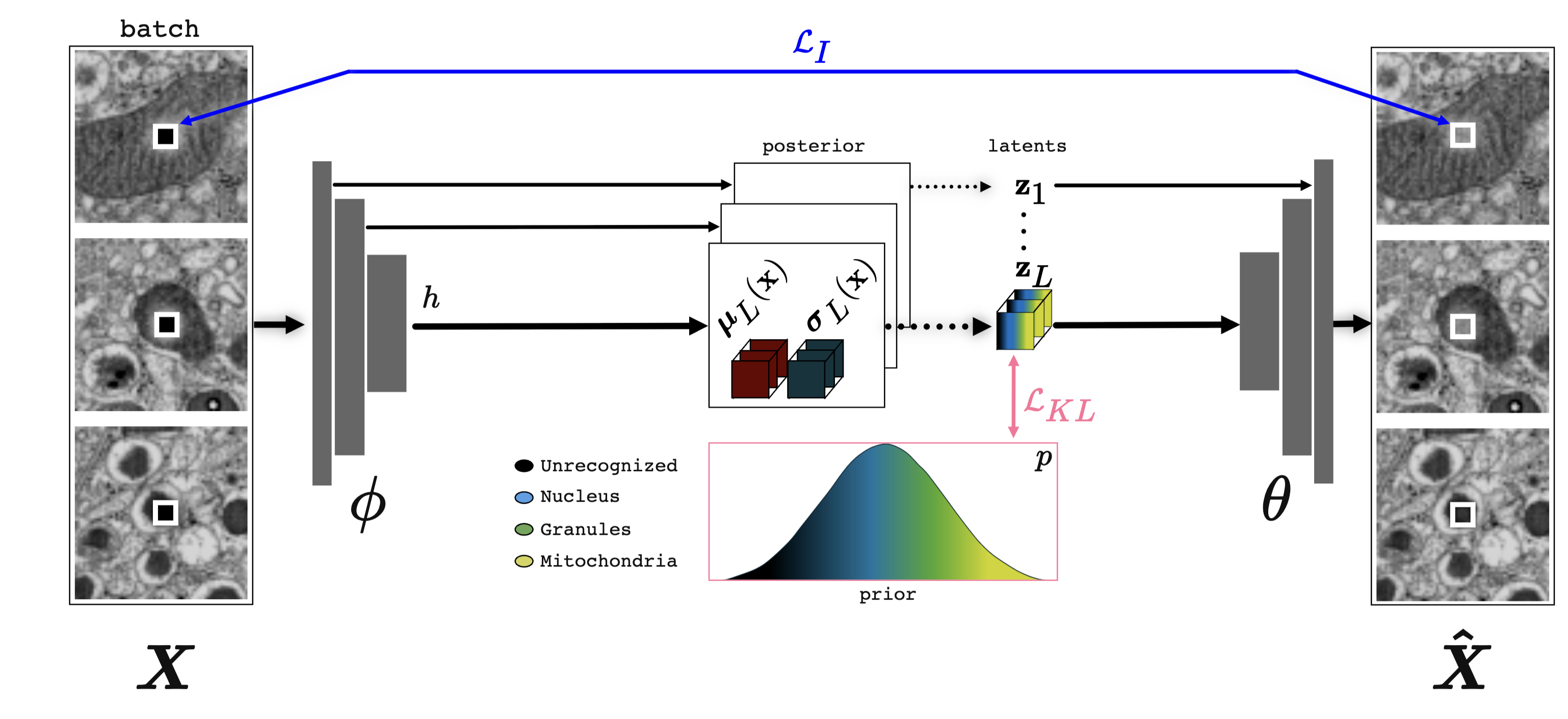}
    \caption{The overall pipeline of Vanilla HVAE in Table~\ref{tab:baseline} (first row in Table~\ref{tab:loss}), which is trained on an inpainting task (of the center-region masked inputs). $\phi$ and $\theta$ are encoder and decoder of the network, respectively. Dotted arrows show sampling from a distribution. $h$ is an intermediate feature embedding of input $\bm{x}$ coming from the encoder $\phi$ and it is posterior distribution's parameters which is divided into two chunks shown as $\bm{\mu}_{L}$ and $\bm\sigma_{L}$. $\bm{z}_L$ is a sample from $\mathcal{N}(\bm\mu_{L}(\bm{x}),\bm\sigma_{L}^2(\bm{x)})$. For $\mathcal{L}_{I}$ inpainting loss and $\mathcal{L}_{KL}$ refer to Equations~\ref{eq:mse_inpainting_loss} and~\ref{eq:hvae_kl} respectively.}
    \label{fig:base}
\end{figure}
}
\newcommand\figCl{
\begin{figure}[hbt]
    \centering
    \includegraphics[width=0.99\textwidth]{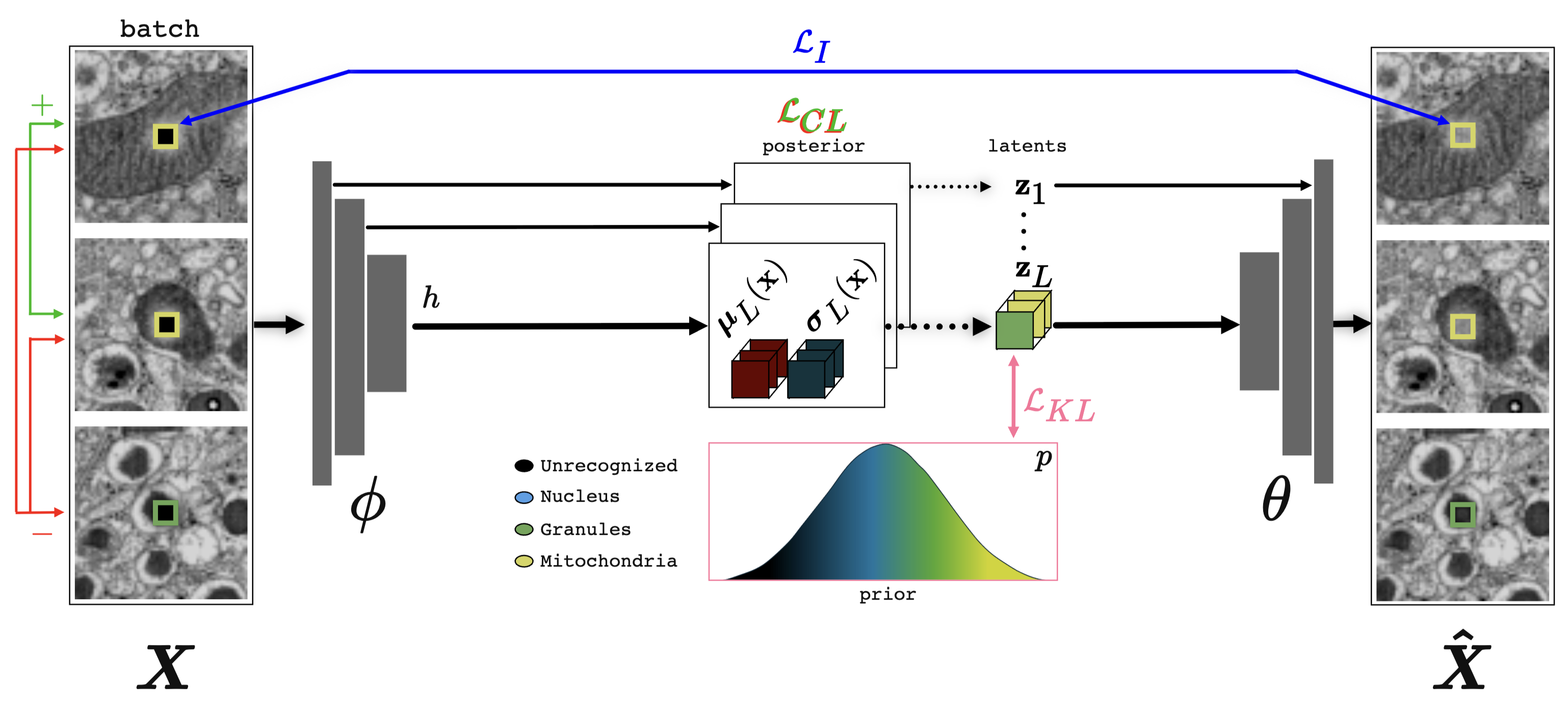} 
    \caption{The overall pipeline of Vanilla HVAE with only CL added in the pipeline in the second row in Table~\ref{tab:loss}, which is trained on an inpainting task (of the center-region masked inputs). Green and red arrows are showing positive and negative pair respectively, in a batch. $\phi$ and $\theta$ are encoder and decoder of the network, respectively. Dotted lines show sampling from a distribution. $h$ is an intermediate feature embedding of input $\bm{x}$ coming from the encoder $\phi$ and it is posterior distribution's parameters which is divided into two chunks shown as $\bm{\mu}_{L}$ and $\bm\sigma_{L}$. $\bm{z}_L$ is a sample from $\mathcal{N}(\bm\mu_{L}(\bm{x}),\bm\sigma_{L}^2(\bm{x)})$. For $\mathcal{L}_{I}$ inpainting loss, $\mathcal{L}_{CL}$ contrastive loss and $\mathcal{L}_{KL}$ refer to Equations~\ref{eq:mse_inpainting_loss},~\ref{eq:cl} and~\ref{eq:hvae_kl} respectively.}
    \label{fig:cl}
\end{figure}
}

\newcommand\figSegModels{
\begin{figure*}
    \centering
    \begin{minipage}{0.9\textwidth}
        \makebox[\textwidth][c]{
            \begin{minipage}{0.33\textwidth}
        \includegraphics[width=\textwidth]{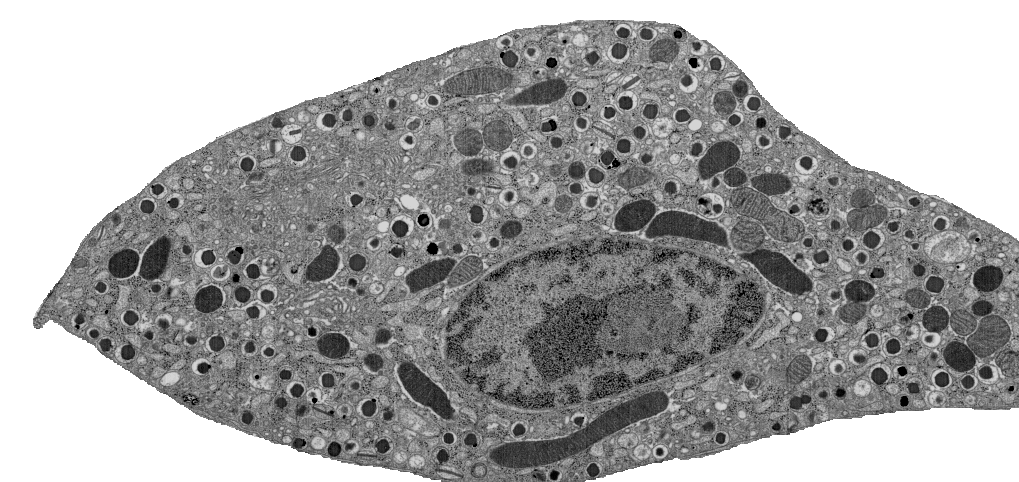}
        \par \centering Image
        \end{minipage}%
        \begin{minipage}{0.33\textwidth}
            \includegraphics[width=\textwidth]{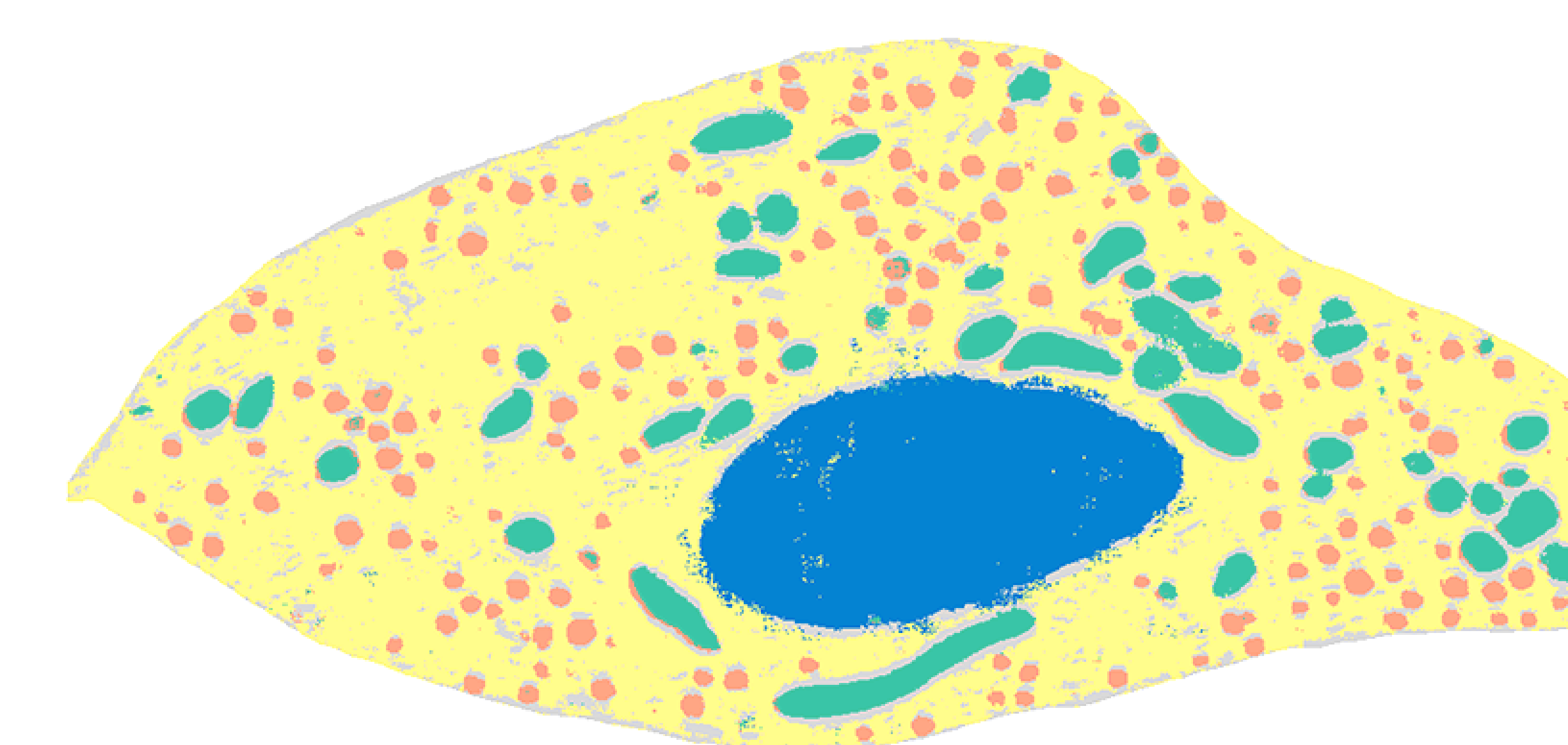}
            \par \centering MAESTER
        \end{minipage}%
        \begin{minipage}{0.33\textwidth}
            \includegraphics[width=\textwidth]{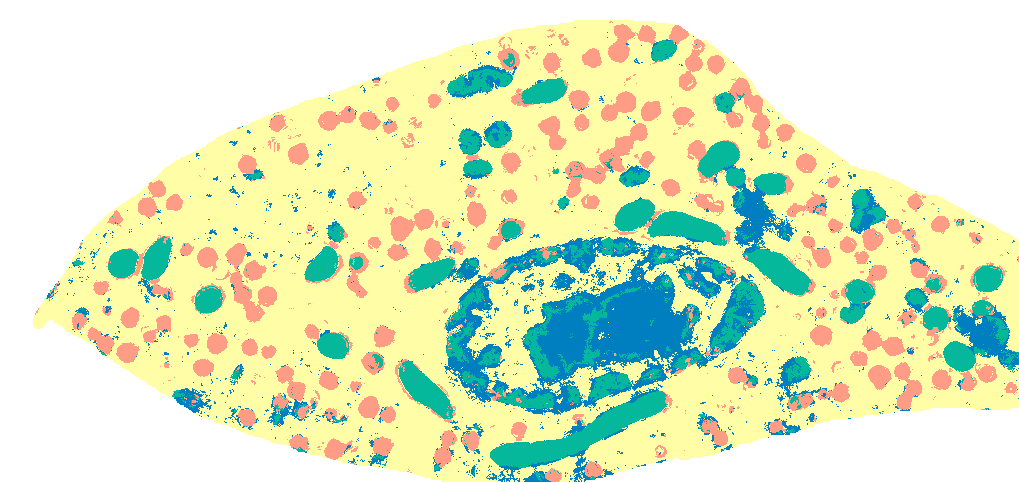}
            \par \centering Labkit
        \end{minipage}
        }
        \makebox[\textwidth][c]{
            \begin{minipage}{0.33\textwidth}
        \includegraphics[width=\textwidth]{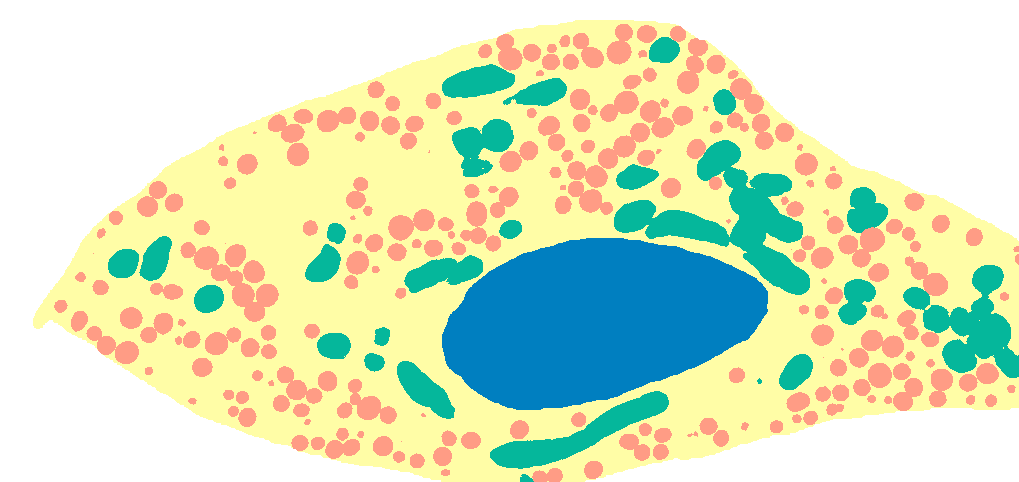}
        \par \centering GT Labels
        \end{minipage}%
        \begin{minipage}{0.33\textwidth}
            \includegraphics[width=\textwidth]{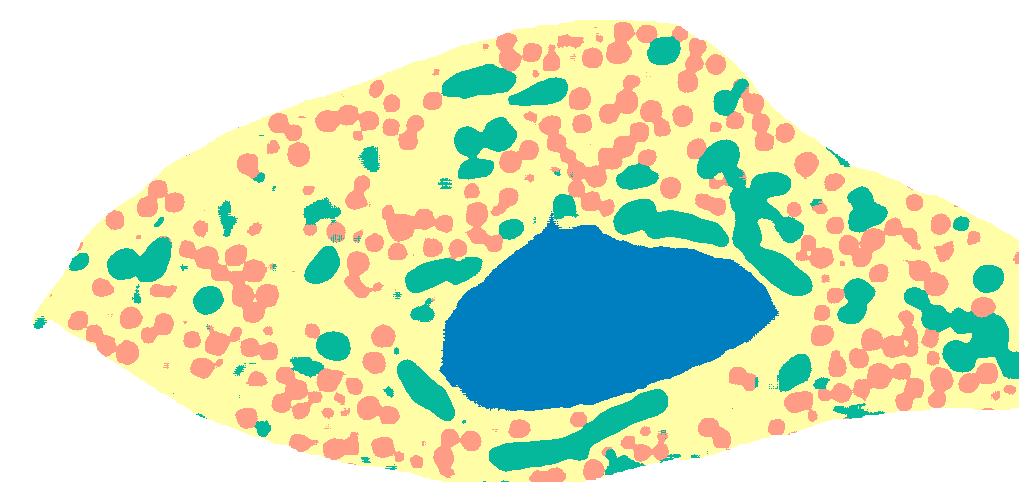}
            \par \centering \textbf{\epsSeg (ours)}
        \end{minipage}%
        \begin{minipage}{0.33\textwidth}
            \includegraphics[width=\textwidth]{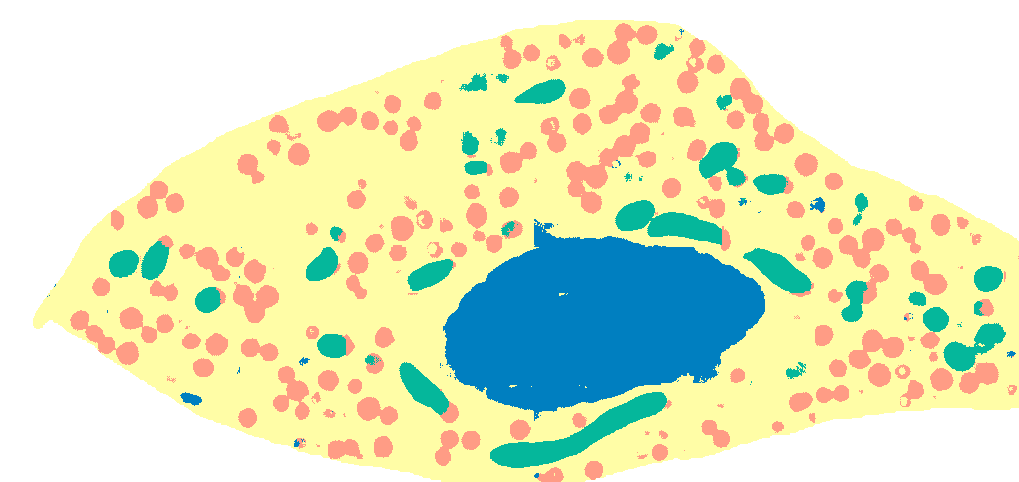}  
            \par \centering U-Net
        \end{minipage}
        }
    \end{minipage}%
    \begin{minipage}{0.10\textwidth}
        \vspace{1.4em}
        \includegraphics[width=\textwidth]{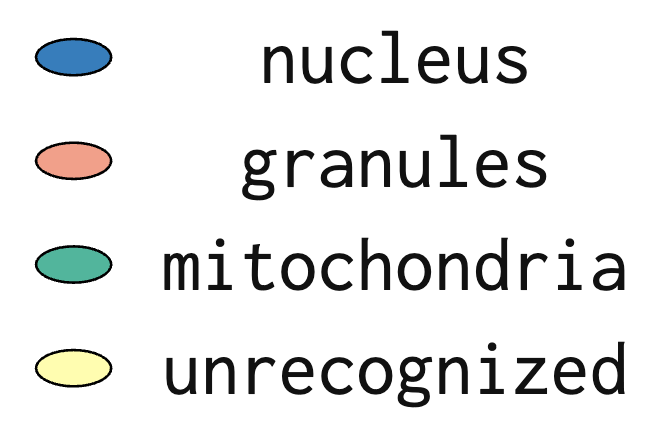}
    \end{minipage}
    \caption{Qualitative segmentation result on part of the test image stack (here we show section $627$ of $high\_c4$ of the ``BetaSeg'' dataset~\cite{Muller2021-mg}).}
    \label{fig:seg_models}
\end{figure*}
}

\newcommand\figSegModelsFull{
\begin{figure*}
    \centering
    \begin{minipage}{0.5\textwidth}
        \includegraphics[width=\textwidth]{img/beta/img.png}
        \par \centering Image
    \end{minipage}%
    \begin{minipage}{0.5\textwidth}
        \includegraphics[width=\textwidth]{img/beta/maester.png}
        \par \centering MAESTER
    \end{minipage}%

    \begin{minipage}{0.5\textwidth}
        \includegraphics[width=\textwidth]{img/beta/gt.png}
        \par \centering GT Labels
    \end{minipage}%
    \begin{minipage}{0.5\textwidth}
        \includegraphics[width=\textwidth]{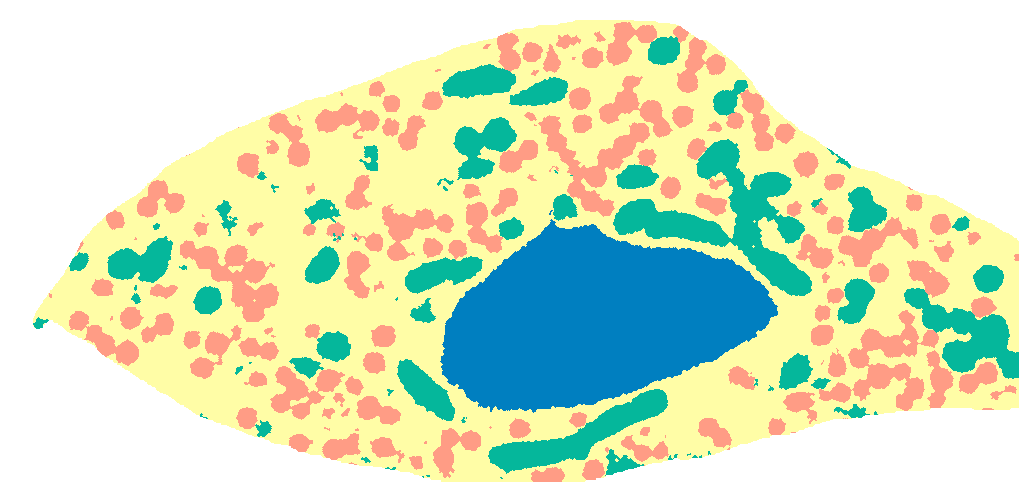}
        \par \centering \textbf{\epsSeg ($+\mathcal{L}_H$)}
    \end{minipage}%

    \begin{minipage}{0.49\textwidth}
        \includegraphics[width=\textwidth]{img/beta/labkit_c1.png}
        \par \centering Labkit
    \end{minipage}
    \begin{minipage}{0.49\textwidth}
        \includegraphics[width=\textwidth]{img/beta/unet_sparse.png}  
        \par \centering U-Net
    \end{minipage}

    \begin{minipage}{0.49\textwidth}
        \includegraphics[width=\textwidth]{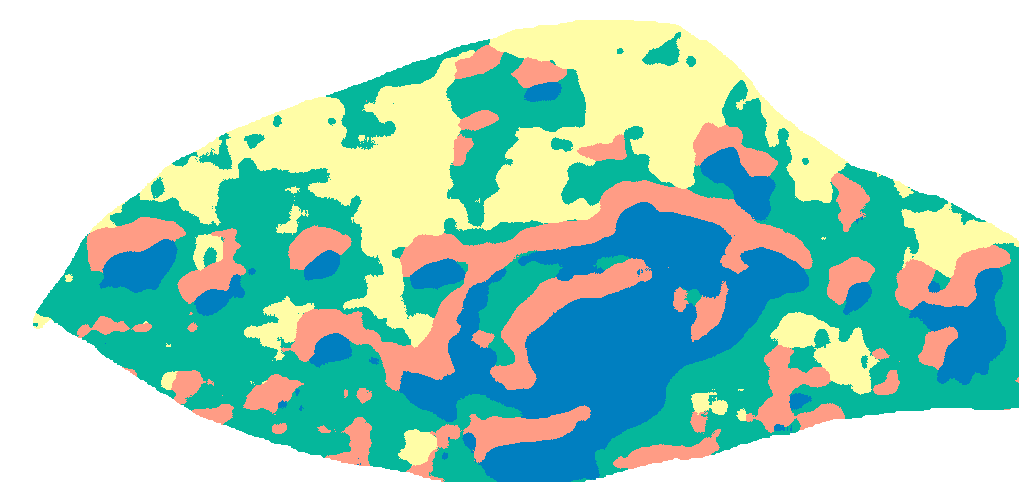}
        \par \centering Vanilla HVAE
    \end{minipage}
    \begin{minipage}{0.49\textwidth}
        \includegraphics[width=\textwidth]{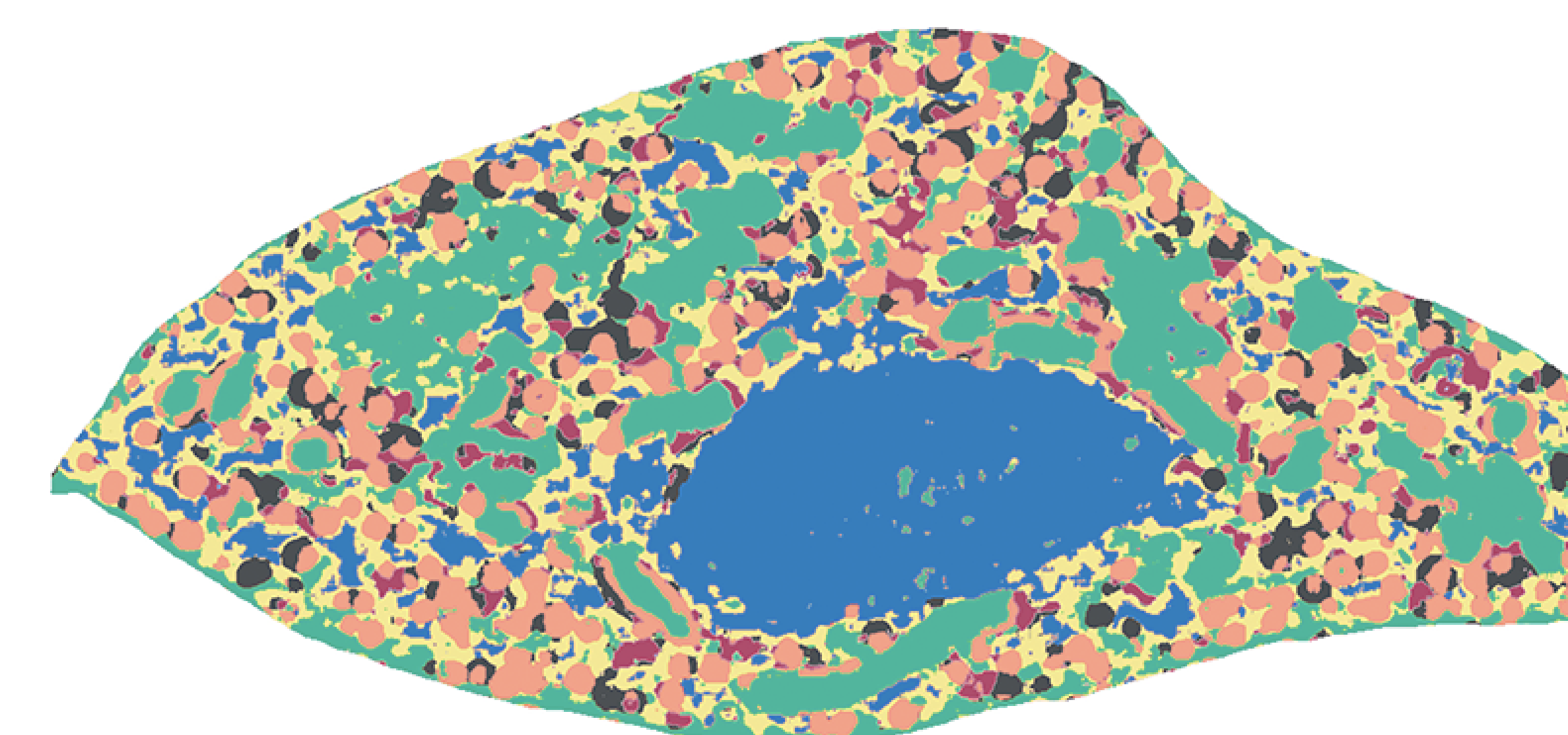}
        \par \centering Han et al.
    \end{minipage}

    \begin{minipage}{0.15\textwidth}
        \includegraphics[width=\textwidth]{img/label1.png}
    \end{minipage}
    
    \caption{Qualitative segmentation result on part of the test image stack (section $627$ of $high\_c4$ in ``BetaSeg'' dataset).}
    \label{fig:seg_models_full}
\end{figure*}
}

\newcommand\figSegModelsZerial{
\begin{figure*}[h]
    \centering
    \begin{minipage}{0.9\textwidth}
        \makebox[\textwidth][c]{
            \makebox[0pt][r]{\textbf{(a)}\hspace{0.5em}}%
            \begin{minipage}{0.23\textwidth}
                \includegraphics[width=\textwidth]{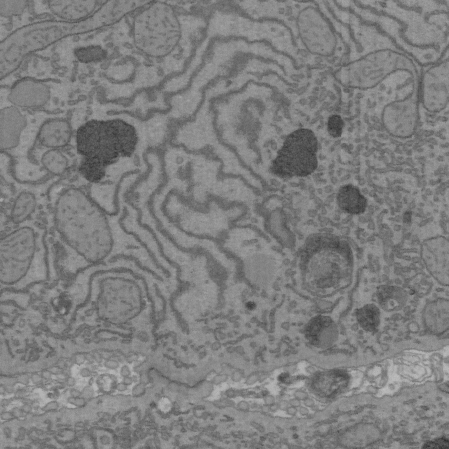}
            \end{minipage}
            \vrule
            \hspace{0.001\textwidth}
            \begin{minipage}{0.23\textwidth}
                \includegraphics[width=\textwidth]{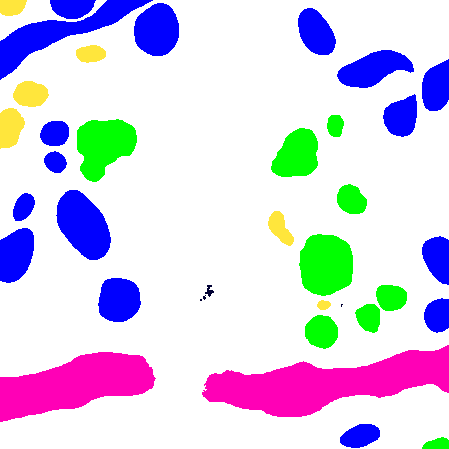} 
            \end{minipage}
            \vrule
            \hspace{0.001\textwidth}
            \begin{minipage}{0.23\textwidth}
                \includegraphics[width=\textwidth]{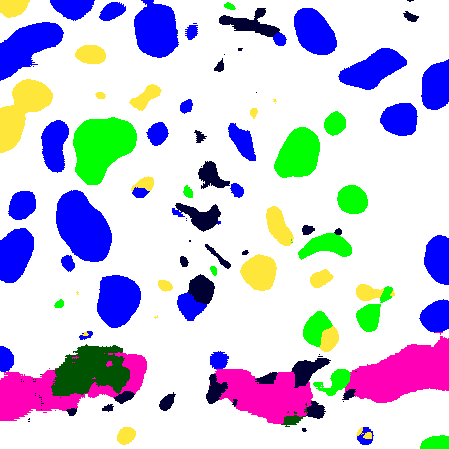}
            \end{minipage}
            \vrule
            \hspace{0.001\textwidth}
            \begin{minipage}{0.23\textwidth}
                \includegraphics[width=\textwidth]{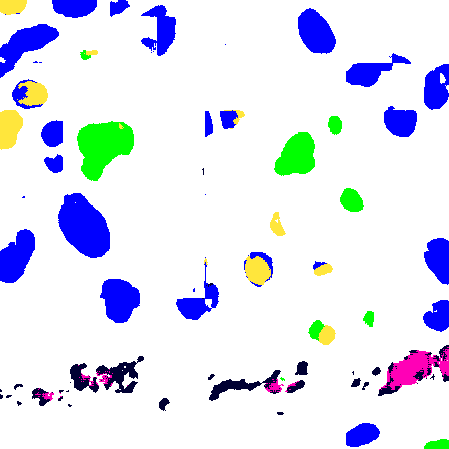}
            \end{minipage}
        }
        \makebox[\textwidth][c]{
            \makebox[0pt][r]{\textbf{(b)}\hspace{0.5em}}%
            \begin{minipage}{0.23\textwidth}
                \includegraphics[width=\textwidth]{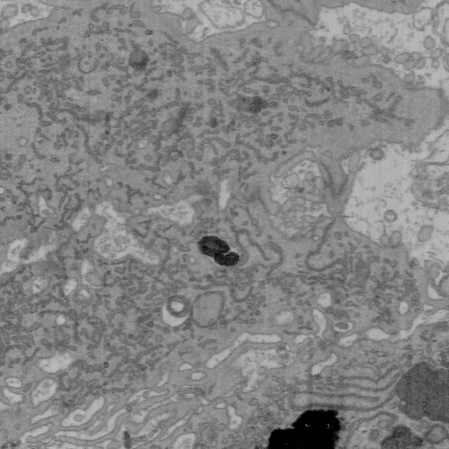}
                \par \centering Image
            \end{minipage}
            \vrule
            \hspace{0.001\textwidth}
            \begin{minipage}{0.23\textwidth}
                \includegraphics[width=\textwidth]{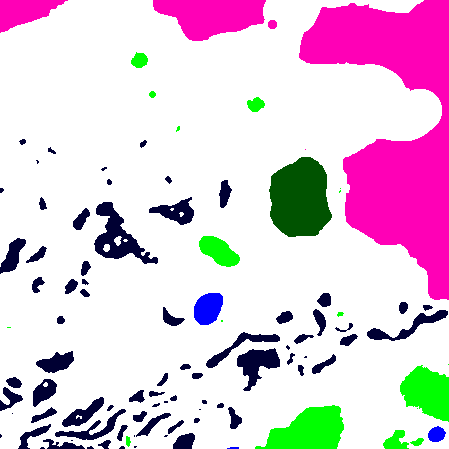}
                \par \centering GT Labels
            \end{minipage}
            \vrule
            \hspace{0.001\textwidth}
            \begin{minipage}{0.23\textwidth}
                \includegraphics[width=\textwidth]{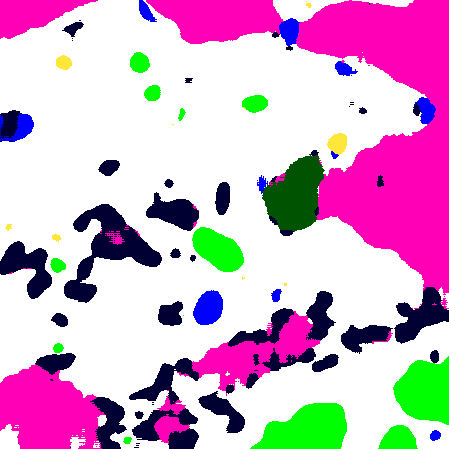}  
                \par \centering \textbf{\epsSeg (ours)}
            \end{minipage}
            \vrule
            \hspace{0.001\textwidth}
            \begin{minipage}{0.23\textwidth}
                \includegraphics[width=\textwidth]{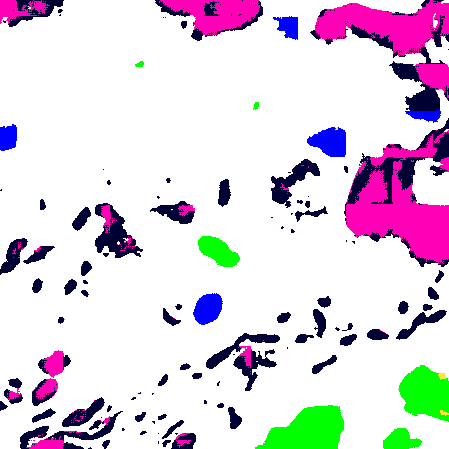}  
                \par \centering U-Net
            \end{minipage}
        }
    \end{minipage}%
    \begin{minipage}{0.09\textwidth}
        \includegraphics[width=\textwidth]{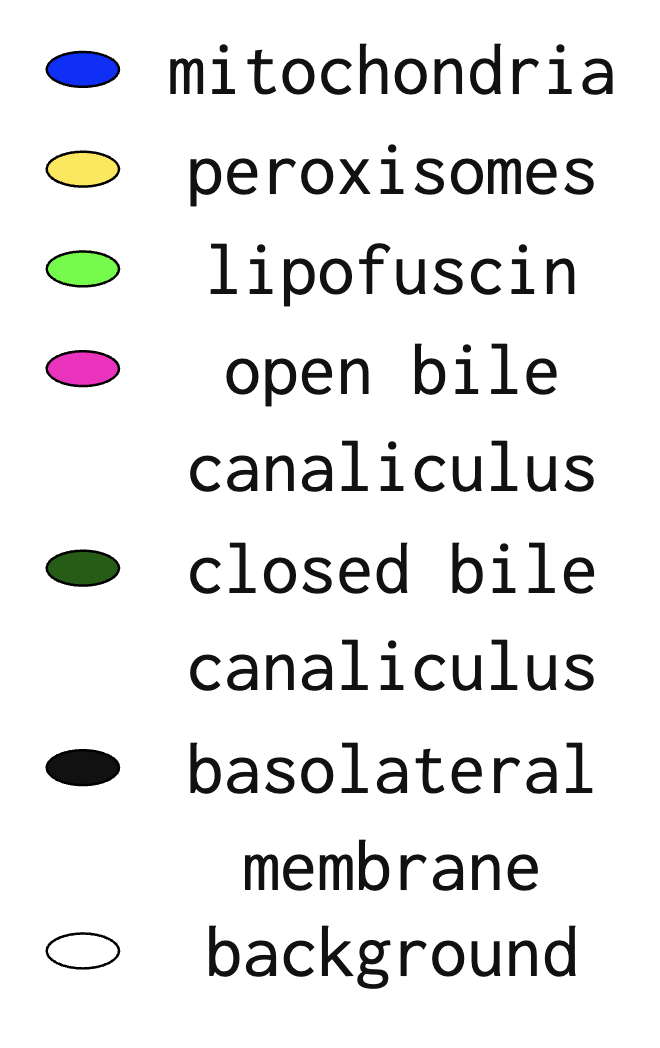}
    \end{minipage}
    \caption{Qualitative segmentation result on two crops of the whole 3D volume. (a) and (b) are section $80$ and $26$ of crop$00$ and crop$10$ in ``liver FIBSEM'' dataset respectively. The U-Net is sparsely-supervised (for the fully-supervised U-Net result, see Figure~\ref{fig:seg_models_zerial_full}).}
    \label{fig:seg_models_zerial}
\end{figure*}
}

\newcommand\figSegModelsZerialFull{
\begin{figure*}
    \centering
    \begin{minipage}{\textwidth}
        \makebox[\textwidth][c]{
            \makebox[0pt][r]{\textbf{(a)}\hspace{0.5em}}\vspace{2em}%
            \begin{minipage}[t]{0.3\textwidth}
                \includegraphics[width=\textwidth]{img/liver/img1.png}
                \par \centering Image
                \hspace{5em}
                \includegraphics[width=0.6\textwidth]{img/label2.png}
            \end{minipage}
            
            \begin{minipage}[t]{0.3\textwidth}
                \includegraphics[width=\textwidth]{img/liver/gt1.png}
                \par \centering GT Labels
                \includegraphics[width=\textwidth]{img/liver/sparse_unet1.png}
                \par \centering U-Net (sparse)
            \end{minipage}

            \begin{minipage}[t]{0.3\textwidth}
                \includegraphics[width=\textwidth]{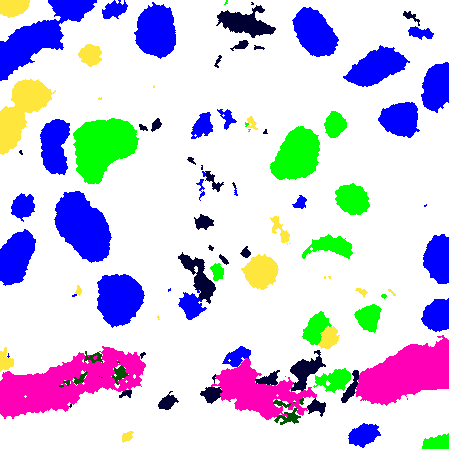}
                \par \centering \textbf{\epsSeg ($+\mathcal{L}_H$)}
                \includegraphics[width=\textwidth]{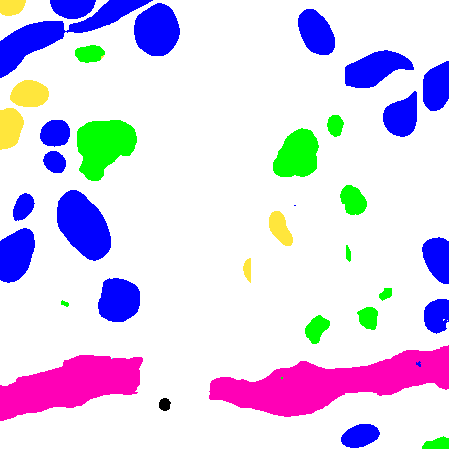}
                \par \centering U-Net (full)
            \end{minipage}
        }
    \end{minipage}
    \hrule
    \hspace{0.1\textwidth}
    \begin{minipage}{\textwidth}
        \makebox[\textwidth][c]{
            \makebox[0pt][r]{\textbf{(b)}\hspace{0.5em}}\vspace{2em}%
            \begin{minipage}[t]{0.3\textwidth}
                \includegraphics[width=\textwidth]{img/liver/img2.png}
                \par \centering Image
                \hspace{5em}
                \includegraphics[width=0.6\textwidth]{img/label2.png}
            \end{minipage}
            
            \begin{minipage}[t]{0.3\textwidth}
                \includegraphics[width=\textwidth]{img/liver/gt2.png}
                \par \centering GT Labels
                \includegraphics[width=\textwidth]{img/liver/sparse_unet2.png}
                \par \centering U-Net (sparse)
            \end{minipage}

            \begin{minipage}[t]{0.3\textwidth}
                \includegraphics[width=\textwidth]{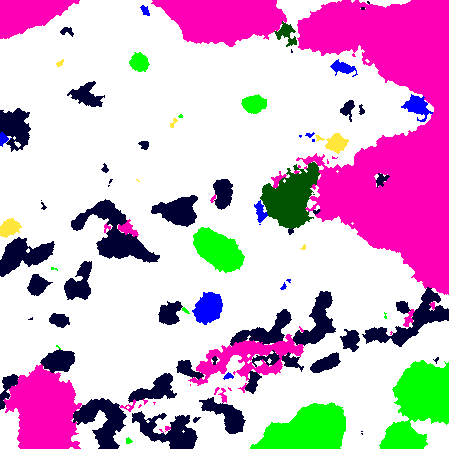}
                \par \centering \textbf{\epsSeg ($+\mathcal{L}_H$)}
                \includegraphics[width=\textwidth]{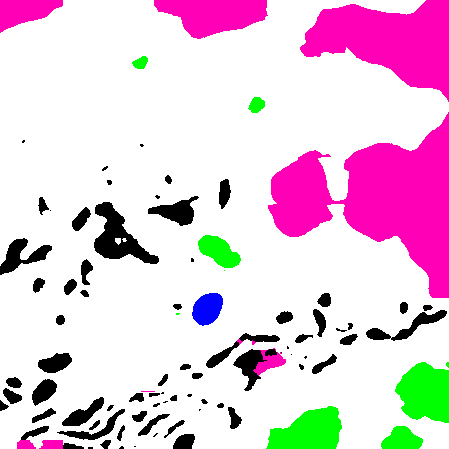}
                \par \centering U-Net (full)
            \end{minipage}
        }
    \end{minipage}
    \caption{Qualitative segmentation result on two crops of the whole 3D volume. (a) and (b) are section $80$ and $26$ of crop$00$ and crop$10$ in ``liver FIBSEM'' dataset respectively. U-Net (sparse) and (full) is sparsely-supervised and fully-supervised respectively.}
    \label{fig:seg_models_zerial_full}
\end{figure*}
}

\newcommand\figFlu{
\begin{figure*}
    \centering
    \begin{minipage}{0.23\textwidth}
        \includegraphics[width=\textwidth]{img/flu/cell.png}
        \par \centering Cytoplasm
    \end{minipage}
    \vrule
    \hspace{0.001\textwidth}
    \begin{minipage}{0.23\textwidth}
        \includegraphics[width=\textwidth]{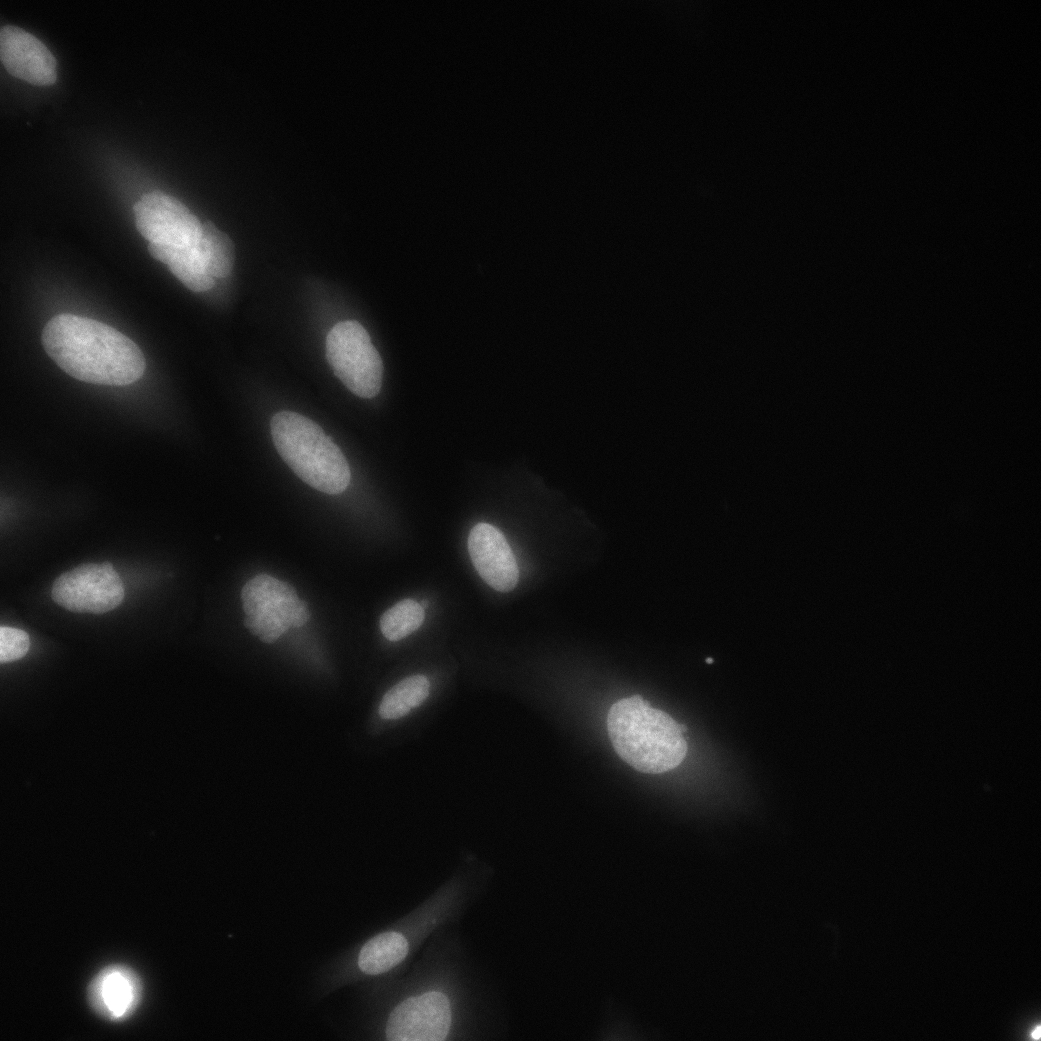} 
        \par \centering Nuclei
    \end{minipage}
    \vrule
    \hspace{0.001\textwidth}
    \begin{minipage}{0.23\textwidth}
        \includegraphics[width=\textwidth]{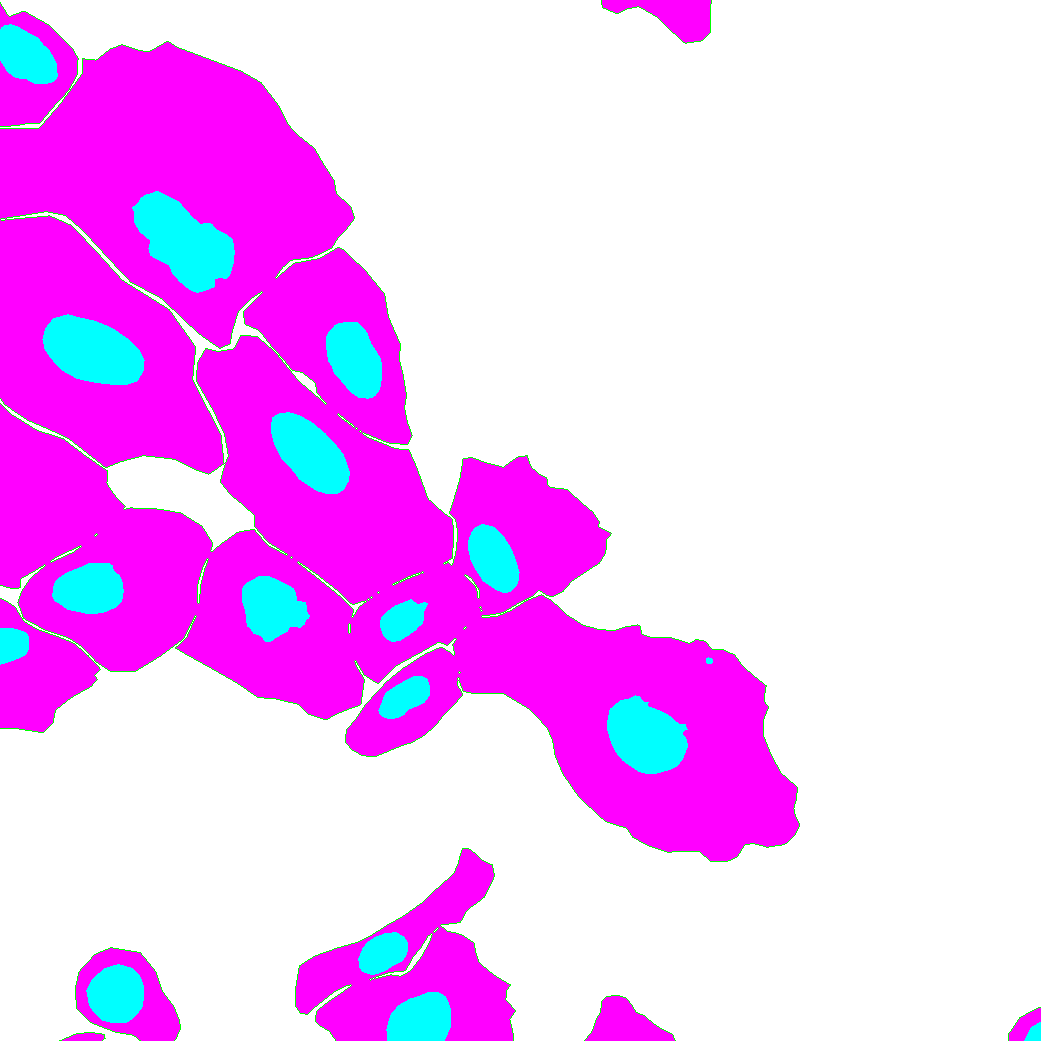}
        \par \centering GT Labels
    \end{minipage}
    \vrule
    \hspace{0.001\textwidth}
    \begin{minipage}{0.23\textwidth}
        \includegraphics[width=\textwidth]{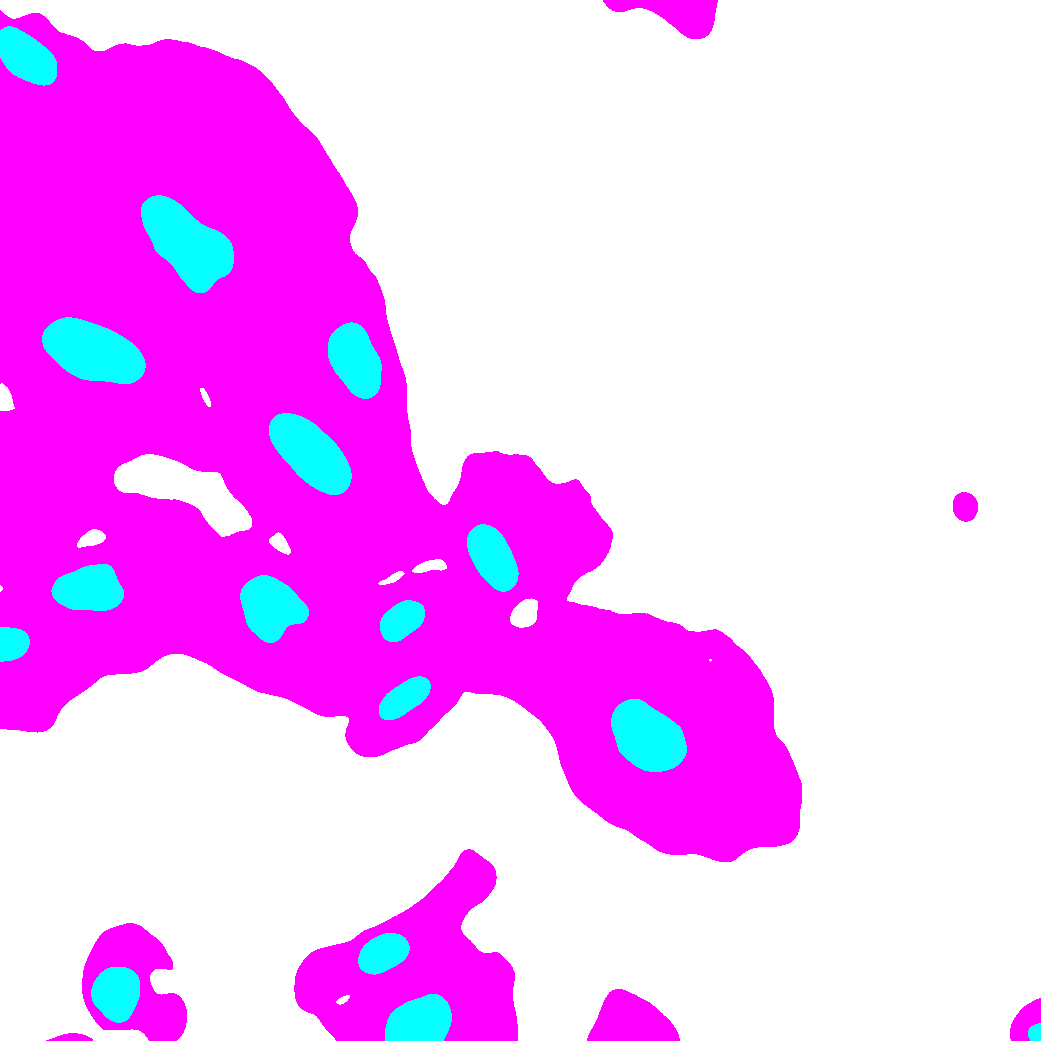}
        \par \centering \epsSeg
    \end{minipage}
    \caption{Qualitative results on a representative 2-channel image from the overlapping subset of the ``Aitslab-bioimaging1'' and ``Aitslab-bioimaging2'' datasets. The first two panels show the fluorescence microscopy channels: EGFP-Galectin-3-labeled cytoplasm (left) and Hoechst 33342-stained nuclei (center-left). The center-right panel (GT) displays the ground truth semantic segmentation with nuclei (cyan) and cytoplasm (magenta). The rightmost panel (\epsSeg) shows the prediction from our method.}
    \label{fig:seg_models_flu}
\end{figure*}
}

%% file: sec/0_abstract.tex
\begin{abstract}
Semantic segmentation of electron microscopy~(\EM) images of biological samples remains a challenge in the life sciences.
\EM data captures details of biological structures, sometimes with such complexity that even human observers can find it overwhelming.
We introduce \epsSeg, a method based on hierarchical variational autoencoders (\HVAEs), employing center-region masking, sparse label contrastive learning (\CL), a Gaussian mixture model (\GMM) prior, and clustering-free label prediction.
Center-region masking and the inpainting loss encourage the model to learn robust and representative embeddings to distinguish the desired classes, even if training labels are sparse ($0.05$\% of the total image data or less).
For optimal performance, we employ \CL and a \GMM prior to shape the latent space of the \HVAE such that encoded input patches tend to cluster \wrt the semantic classes we wish to distinguish. 
Finally, instead of clustering latent embeddings for semantic segmentation, we propose a \MLP semantic segmentation head to directly predict class labels from latent embeddings.
We show empirical results of \epsSeg and baseline methods on $2$ dense \EM datasets of biological tissues and demonstrate the applicability of our method also on fluorescence microscopy data. 
Our results show that \epsSeg is capable of achieving competitive sparsely-supervised segmentation results on complex biological image data, even if only limited amounts of training labels are available.
Code available at https://github.com/juglab/eps-Seg.
\end{abstract}

%% file: sec/1_intro.tex
\section{Introduction}
\label{sec:intro}

Electron Microscopy~(\EM) comes in multiple flavors and is without doubt the tool of choice for high-resolution investigations of biological samples~\cite{Drobne2013}.
Today, microscopists can capture fine cellular structures at nanometer resolution~\cite{Muller2021-mg,Anusha2022_Segmentation}.
Although this opens unprecedented possibilities for studying the very fabric of life, it also means that such microscopes are producing an unfathomable amount of raw image data that then are available to be analyzed~\cite{Xu2021-ew}.

A key module of nearly every analysis pipeline is the segmentation step, where specific structures of interest must be found in the entire body of captured image data. 
Performing this step manually, is typically not feasible as it takes an impossibly long time~\cite{Heinrich2021-la,Xu2021-ew,Muller2021-mg}.
Unfortunately, even semantic segmentation of \EM data of biological samples remains a challenge~\cite{Anusha2022_Segmentation, Treder2022ApplicationsOD}.

Ideally, methods for segmenting \EM data should
$(i)$~lead to sufficiently good segmentation results for the downstream analysis tasks at hand with as few training labels as possible,
$(ii)$~generalize well to different imaging conditions and image tissue types and/or be able to fine-tune on moderate amounts of new training data~\cite{CEM500K},
$(iii)$~be able to benefit from sparse labeled data via supervised contrastive learning approaches, and if possible
$(iv)$~operate on a hierarchy of spatial scales to distinguish objects not only by either detailed textures or larger scale shapes, but both. 

With this in mind, we introduce \epsSeg, a novel and sparsely supervised semantic segmentation framework for \EM images that
reduces the `hunger' for labeled data by using a powerful hierarchical \VAE (\HVAE)~\cite{lvae,very_deep} with a \GMM prior instead of a regular Gaussian one.
Furthermore, our method uses center-region inpainting and contrastive learning to enhance feature consistency and segmentation robustness, even when training data is scarce.
Hence, \epsSeg learns structured latent space representations with effective feature separation for the semantic classes of interest. 
Once such features are learned, they can be clustered to obtain meaningful semantic segmentations.
However, since this process is computationally intensive, we integrate a dedicated semantic segmentation head that directly produces segmentation labels, thereby improving both accuracy and runtime.

%% file: sec/2_related_work.tex

\section{Related Work}
\label{sec:related_work}
\miniheadline{Sparse Supervision}
Deep learning has transformed microscopy image segmentation.
The \UNet~\cite{unet} has long been a standard architecture, achieving strong results when trained in a fully supervised setting.
However, such approaches rely on dense annotations, which are costly and time-consuming to obtain.
At the other extreme, self-supervised methods such as MAESTER~\cite{maester} learn directly from raw data without labels, offering excellent scalability but typically at the cost of reduced segmentation accuracy compared to fully supervised approaches.
Between these extremes lies a growing body of work on sparse or weak supervision, which seeks to achieve label efficiency while maintaining good performance.
We aim to surpass self-supervised methods in accuracy while requiring only a fraction of the annotations needed by fully supervised methods.
Comprehensive reviews on segmentation methods in large-scale EM with deep learning are available~\cite{Anusha2022_Segmentation}, with representative examples including slice-wise pseudo-label propagation for neuronal membranes (4S)~\cite{takaya} , or domain adaptation variants of \UNet designed for limited-annotation settings~\cite{bermudez}.

\miniheadline{Hierarchical Variational Autoencoders}
Hierarchical architectures, like \HVAEs~\cite{lvae, very_deep, nvae, child, hdn}, appear to be an interesting choice for segmenting biological microscopy data.
Based on variational autoencoders~\cite{Kingma2014}, these powerful models learn a full approximate posterior, but are limited by the typically used Gaussian prior, making us wonder if a Gaussian mixture would not be a more suitable choice for the semantic segmentation task at hand.
While the above-mentioned methods pursue label efficiency through different strategies, they do not explicitly enforce semantically disentangled latent representations.
In contrast, we explicitly enforce semantically disentangled latent representations by combining a \GMM prior with contrastive learning, ensuring that each latent component aligns with a distinct object class.
This motivates our focus on \HVAEs, which progressively encode features from fine to coarse across network layers.
As higher-level semantic structure emerges in deeper layers, the latent space can be disentangled and aligned with semantic classes, enabling efficient segmentation and downstream biological analysis.

\miniheadline{Gaussian Mixture Models (GMMs)}
GMMs have been extensively used to model multimodal distributions and are a key component for many clustering methods~\cite{clustergmvae, segma, clusterae, gmvae}.
Many approaches integrate GMMs within autoencoder-based architectures, either explicitly as a clustering module~\cite{clusterae} or by enforcing multimodal latent structure through a GMM prior~\cite{clustergmvae, gmvae}.
In \VAEs, GMM priors enable structured latent spaces where each mixture component represents a distinct cluster or class~\cite{gmvae, clustergmvae}.
Some methods employ direct optimization of GMM objectives alongside autoencoders~\cite{clusterae}, while others leverage categorical latent variables within GMVAE frameworks, using discrete reparameterization techniques such as the Gumbel-Softmax~\cite{gumbel} relaxation to improve scalability~\cite{clustergmvae}.
These techniques effectively combine deep generative models with Gaussian mixture priors, enhancing unsupervised representation learning and clustering performance in high-dimensional data spaces.

\miniheadline{Contrastive Learning (CL)}
CL has gained attention for its ability to refine feature representations by maximizing similarities between related samples and minimizing them between unrelated ones.
Methods like SimCLR~\cite{chen_cl} and MoCo~\cite{moco} demonstrated their effectiveness in many applications.
In the context of EM segmentation, CL enables better alignment of latent representations with subcellular structures.
We will use CL to ensure that each GMM component corresponds to a distinct semantic class, not just in the highest level of the hierarchy we learn.

Next, we present our proposed method, which integrates hierarchical variational autoencoders with GMM-based priors and contrastive learning to achieve accurate and label-efficient EM segmentation.

%% file: sec/3_method.tex
\section{Methods}
\label{sec:method}
\figMain
The method we propose is based on a Hierarchical \VAE (\HVAE) backbone similar to the ones described in~\cite{lvae,hdn}.
We modify the standard \HVAE setup by 
$(i)$~using a Gaussian mixture model (\GMM) instead of the default Gaussian, so every semantic class we want to distinguish has its own predetermined Gaussian region, and by
$(ii)$~adding a contrastive loss (\CL), we further ensure that latent encodings are grouped by their semantic similarity through all hierarchy levels. 

As the basis for our work, we used the openly available \HVAE backbone of Hierarchical DivNoising (HDN)~\cite{hdn}. 
\HVAEs, as introduced elsewhere~\cite{lvae,nvae,very_deep,hdn}, consist of a bottom-up path (encoder) and a top-down path (decoder) with trainable parameters $\phi$ and $\theta$, respectively.
The encoder extracts features from a given input $\bm{x}$ at progressively coarser scales, creating a hierarchical latent encoding $\bm{z}$ that splits into sub-spaces $\bm{z}_i, i=1\dots L$, with $L$ being the number of hierarchy levels, or latent layers, in the \HVAE.
The decoder network in regular \HVAEs reconstructs $\bm{x}$, starting from the topmost latent variables $\bm{z}_L$.
Here, we first switch from reconstructing $\bm{x}$ to inpainting a masked central region in $\bm{x}$, as described next.

\miniheadline{Autoencoding vs.\ Inpainting}
In contrast to regular \VAEs and \HVAEs that use a reconstruction loss on full input patches $\bm{x}$, we are using masked autoencoding instead~\cite{cmae}.
Since our aim is to learn semantic features that can be used for pixel-level semantic segmentation, the zero-masking we employed asks the network to only reconstruct the masked region, effectively learning features that best represent the masked semantic class.
We conducted experiments with masked regions of various sizes and have always ensured that all masked pixels were from the same semantic class, see Table~\ref{tab:mask}.

The model is trained to reconstruct the masked center pixel(s) using an MSE-based inpainting loss on $\bm{X}$, a training batch of inputs, of size $B$, as 
\begin{equation}
\mathcal{L}_{\text{I}} = \frac{1}{B} \sum_{\bm{x}\in \bm{X}} \left( \bm{x}^{\text{mask}} - \hat{\bm{x}}^{\text{mask}} \right)^2,
\label{eq:mse_inpainting_loss}
\end{equation}
where $\hat{\bm{x}}^{\text{mask}}$ is the inpainted masked region the decoder predicted, and $\bm{x}^{\text{mask}}$ is the mask region of the respective input patch prior to zero-masking.

\miniheadline{HVAEs with Gaussian Priors}
The Gaussian prior of regular \VAEs only applies to the topmost hierarchy level in \HVAEs, where it remains $\mathcal{N}(0, I)$ as depicted in Figure~\ref{fig:base}.

The latent variables $\bm{z}$ of a \HVAE are split into $L$ layers $\bm{z}_i, i\in [1, \dots, L]$ so that
\begin{equation}
    p_{\theta}(\bm{z}) = p_{\theta}(\bm{z}_{L}) \prod_{i=1}^{L-1} p_{\theta}(\bm{z}_{i}|\bm{z}_{i+1}),
    \label{eq:pofz}
\end{equation}
\begin{equation}
    p_{\theta}(\bm{z}_{L}) = \mathcal{N}(\bm{z}_L|\bm{0},\bm{I}),
    \label{eq:pofzL}
\end{equation}
\begin{equation}
    p_{\theta}(\bm{z}_i|\bm{z}_{i+1}) = \mathcal{N}(\bm{z}_i|\mu_{p,i}(\bm{z}_{i+1}), \sigma^2_{p,i}(\bm{z}_{i+1})) \text{ and}
    \label{eq:pofzi}
\end{equation}
\begin{equation}
    p_{\theta}(\bm{x}|\bm{z}_{1}) = \mathcal{N}(\bm{x}|\mu_{p,0}(\bm{z}_{1}), \sigma^2_{p,0}(\bm{z}_{1})),
    \label{eq:pofx}
\end{equation}
where $\mu_\theta(\bm{z}_{i})$ and $\sigma^2_\theta(\bm{z}_{i})$ represent the mean and the variance of the latent encoding, parameterized by $\theta$.

For each layer $i$, the approximate posterior $q_\phi(\bm{z}_i | \bm{x}, \bm{z}_{<i})$, computed by the encoder, is defined as
\begin{equation}
q_\phi(\bm{z}_i | \bm{x}, \bm{z}_{<i}) = \mathcal{N}(\bm{z}_i; \mu_\phi(\bm{x}, \bm{z}_{<i}), \sigma^2_\phi(\bm{x}, \bm{z}_{<i})),
\end{equation}
where $\mu_\phi(\bm{x}, \bm{z}_{<i})$ and $\sigma_\phi(\bm{x}, \bm{z}_{<i})$ are functions parameterized by $\phi$, and are the mean and variance conditioned on the input $\bm{x}$ and the latent variables from lower layers $j<i$, denoted by $\bm{z}_{<i}$.

The KL divergence term for each layer in the Evidence Lower Bound (ELBO) is 
\begin{equation} 
\mathbb{E}_{q_\phi(\bm{z}_{>i} | \bm{x})} 
\left[
    \text{KL}\left(
        q_\phi(\bm{z}_i | \bm{x}, \bm{z}_{<i}) \,\|\, 
        p_\theta(\bm{z}_i | \bm{z}_{i+1})
    \right)
\right],
\label{eq:hvae_kl}
\end{equation}
where $\bm{z}_{>i}$ are all $\bm{z}_j$ for $j>i$.

\miniheadline{HVAEs with a GMM Prior}
When replacing the topmost prior $p_\theta(\bm{z}_L)$ in an \HVAE with a Gaussian mixture model~(\GMM), the prior becomes a weighted sum of Gaussians
\begin{equation}
p_\theta(\bm{z}_L) = \sum_{c=1}^{C} \pi_c \, \mathcal{N}(\bm{z}_L; \mu_c, \sigma^2_c),
\label{eq:gmmprior}
\end{equation}
where $C$ is the total number of Gaussian components and also the number of semantic classes we want to distinguish, $\pi_c$ are the mixing coefficients of the \GMM with $\sum_{c=1}^{C} \pi_c = 1$, and $\mathcal{N}(\bm{z}_L; \mu_c, \sigma^2_c)$ is a Gaussian component with mean $\mu_c$ and standard deviation $\sigma_c$.

Note that there is a one-to-one correspondence between Gaussian components of the \GMM and the semantic classes \epsSeg is supposed to distinguish. 
This would ensure that the latent variable follows a categorical distribution over the semantic classes; we ideally want the mixture assignment $\bm{\pi}=(\pi_1,\dots,\pi_C)$ to act as a one-hot vector, \ie one $\pi_c$ should be $1$, and the rest should be $0$.

However, in practice, learning a fully discrete $\bm\pi$ is challenging because the standard \VAE framework with a \GMM prior typically results in soft assignments~\cite{gmvae}.
To encourage hard assignments, one could ($i$) use a Gumbel-Softmax~\cite{gumbel} trick to approximate categorical sampling while maintaining differentiability~\cite{clustergmvae}, ($ii$) introduce an entropy loss to encourage $\pi_c$ values to be closer to either $0$ or $1$.
In our experiments, we used the Gumbel-Softmax during training, while reverting to the standard softmax at inference time.
We also introduced an entropy loss term as a form of self-supervision, which yielded moderate improvements in the Gumbel-Softmax-based results (see Supplementary Material), but did not lead to significant gains \wrt the best-performing softmax configuration.
We therefore report the softmax-based results as our main findings, without the additional training phase using the entropy loss.
In future work, we plan to investigate alternative self-supervision strategies to further enhance the segmentation performance, leveraging the vast amount of available unlabeled data, within the proposed framework.

The approximate posterior for the topmost latent $\bm{z}_L$, can now be expressed as
\begin{equation}
    q_\phi(\bm{z}_L | \bm{x}) = 
    \sum_{l=1}^{C} q_\phi(c=l | \bm{x})\, q_\phi(\bm{z}_L | \bm{x}, c=l),
    \label{eq:q_gmm}
\end{equation}
where $q_\phi(c | \bm{x})$ is the approximate posterior probability of the \GMM component $c$ set to label $l$ given input $\bm{x}$ and 
$q_\phi(\bm{z}_L | \bm{x}, c)$ is the topmost approximate posterior conditioned on $\bm{x}$ and component $c$. 
We model $q_\phi(\bm{z}_L | \bm{x}, c)$ over all possible labels itself with a Gaussian
\begin{equation}
    q_\phi(\bm{z}_L \mid \bm{x}, c) = \mathcal{N}(\bm{z}_L; \mu_{\bm{L}}(\bm{x}), \sigma_{\bm{L}}(\bm{x})),
    \label{eq:q_gmm_normal}
\end{equation}
by predicting $\mu_{\bm{L}}(\bm{x})$ and $\sigma_{\bm{L}}(\bm{x})$ (see boxes labeled with ``posterior'' in Figure~\ref{fig:main}).
In practice, the parameters $\mu_{\bm{L}}(\bm{x})$ and $\sigma_{\bm{L}}(\bm{x})$ are computed once from the FiLM-conditioned encoder output and are shared across all components $l$.
As a result, the mixture in Equation~\ref{eq:q_gmm} reduces to
\begin{equation}
    \begin{aligned}
    q_\phi(\bm{z}_L | \bm{x}) = 
    \mathcal{N}(\bm{z}_L; \mu_{\bm{L}}(\bm{x}), \sigma_{\bm{L}}(\bm{x})),
    \label{eq:q_condition}
    \end{aligned}
\end{equation}
as depicted in Figure~\ref{fig:main}.
In order to predict $\mu_{\bm{L}}(\bm{x})$ and $\sigma_{\bm{L}}(\bm{x}))$, we must compute the conditional posterior.

\miniheadline{Computing the Conditional Posterior}
In this section, we describe the main backbone of our method leading from a given input patch $\bm{x} \in \bm{X}$ to the computed posteriors $q_\phi=\mathcal{N}(\bm\mu(\bm{x}),\bm\sigma^2(\bm{x}))$.
Figure~\ref{fig:main} illustrates the overall pipeline of \epsSeg.

The encoder, parametrized by $\phi$, processes $\bm{x}$, leading to intermediate features $h$ in the topmost hierarchy level $L$.
These features are then passed through an MLP classifier (rouge box in Figure~\ref{fig:main}), producing a vector of logits $f(h)$ with dimensionality $C$, coinciding with the number of classes \epsSeg is tasked to distinguish.

Instead of directly using $h$ as our posterior distribution parameters, as done in our Vanilla \HVAE baseline, we are using $f(h)$, fed through two additional MLPs, $g_{\gamma}$ and $g_{\beta}$ (see violet boxes in Figure~\ref{fig:main}), to compute parameters, $\gamma$ and $\beta$ such that $\gamma = g_{\gamma} (f(h)) \text{ and } \beta = g_{\beta} (f(h))$.

Those MLPs are mapping logits $f(h)$ into feature-wise scaling and shifting factors.
In this way, the encoded features $h$ are modulated via these FiLM~\cite{film} parameters $\gamma$ and $\beta$ into $h'$ via computing
$
  h' = \gamma \odot h + \beta,
$
where $\odot$ denotes the Hadamard product (element-wise multiplication).
The modulated feature representation $h'$ is then chunked into two parts, $\bm\mu_L(\bm{x})$ and $\bm\sigma_L(\bm{x})$, and used to parameterize the conditional Gaussian posterior in Equation~\ref{eq:q_condition}.

\miniheadline{The Latent Sematic Segmentation Head}
\label{mini:seg}
To avoid computationally costly downstream latent space clustering to perform the semantic segmentation task (as done in~\citet{maester} and~\citet{han} using K-Means clustering), we are introducing a segmentation head tasked to perform the semantic pixel classification tasks directly from the computed logits $f(h)$.

To compute $q_\phi(c|\bm{x})$ of Equation~\ref{eq:q_gmm}, we use a categorical reparameterization trick via Gumbel-Softmax~\cite{gumbel}.

The standard Gumbel-Softmax formula using the class probabilities $\pi_i$ is
\begin{equation}
    y'_i = \frac{\exp((\log \pi_i + g_i)/\tau)}{\sum_{j=1}^{C} \exp((\log \pi_j + g_j)/\tau)},
\end{equation}
where $g_i \sim \text{Gumbel}(0,1)$ are Gumbel noise samples.
Instead of probabilities $\pi_i$, we work with logits $f(h)$ (raw scores before softmax).
The equivalent formula becomes
\begin{equation}
    y'_i = \frac{\exp((f_i(h) + g_i)/\tau)}{\sum_{j=1}^{C} \exp((f_j(h) + g_j)/\tau)}.
\end{equation}

The temperature parameter $\tau$ in the Gumbel-Softmax distribution plays a crucial role in controlling the degree of discreteness in the sampled values.
During training, $\tau$ is often annealed from a higher value to a lower one, gradually transitioning from a smooth approximation to a discrete categorical distribution.

In \epsSeg, we use a typical annealing schedule $\tau = \max(\tau_{\min},\exp(-rt))$, where $r=0.999$ is the decay rate, $\tau_{\min}=0.5$, and $t$ is the training step.
Therefore, Gumbel enables the differentiable sampling of categorical variables, improving gradient estimation, and semi-supervised classification~\cite{gumbel}.

Next, we draw a vector $\bm{y}'$, representing the class assignment (segmentation prediction) for an input patch $\bm{x}^{(i)}$ in the batch $\bm{X}$, by sampling from the Gumbel-Softmax distribution parameterized by logits $f(h)$ with temperature $\tau$.

For input patches $\bm{x}^{(i)}\in \bm{X}$ for which we know the class label $l_i$, we want to ensure that $y'^{(i)}_l\in \bm{y}'^{(i)}$ is the largest entry.
We do so using the cross-entropy loss
\begin{equation}
    \mathcal{L}_{CE} = 
        -\sum_{\bm{x}^{(i)} \in \bm{X}}log \, y'^{(i)}_{l}. 
    \label{eq:ce}
\end{equation}

\miniheadline{Computing the Kullback Leibler Divergence}
As it is commonly done in \VAEs~\cite{Kingma2014}, the KL-divergence term is regularizing the parameters of our encoder, $\phi$, such that the approximate posterior will be close to our prior $p_\theta(\bm{z})$.
In \HVAEs, KL is computed at each hierarchy level.
Changing from a standard Gaussian prior at the highest hierarchy level $L$ to using a \GMM prior, as described earlier in this section, requires us to define a strategy to compute the KL-divergence appropriately.

\citet{gmmkl} address the challenge of efficiently approximating the KL divergence between two GMMs, and~\citet{gmmklul} propose lower and upper bounds to estimate this divergence.
While these approaches can be needed in practical setups~\cite{gmvae, clustergmvae}, we only need to compute the KL divergence between the posterior $q_\phi(\bm{z}_L | \bm{x})$ (Equation~\ref{eq:q_condition}) and the $l$-th GMM component, where $l$ is either the known class label for an input patch $\bm{x}^{(i)}$, or 
$l = \underset{{y'^{(j)}} \in {\bm{y}'^{(j)}}}{\arg\max}\quad{y'^{(j)}}$
for a patch $\bm{x}^{(j)}$ for which we do not have a ground truth class label.

Hence, Equation~\ref{eq:gmmprior} becomes
$
    p_{\theta,c}(\bm{z}_{L})=
        \mathcal{N}(\bm{z}_L; \mu_l, \sigma^2_l),
$
and $\mathcal{L}_{KL}$ is therefore still computed as the divergence between two normal distributions.
The KL loss over all hierarchy levels is therefore 
\begin{equation}
    \resizebox{0.93\textwidth}{!}{
    $\mathcal{L}_{KL} = - (\text{KL}(q_\phi(\bm{z}_1|\bm{x}) \parallel p_\theta(\bm{z}_1|\bm{z}_2)) 
  + \sum_{i=2}^{L-1}\text{KL}(q_\phi(\bm{z}_i|\bm{z}_{i-1})\parallel p_\theta(\bm{z}_i|\bm{z}_{i+1})) 
  + \text{KL}(q_\phi(\bm{z}_L | \bm{z}_{L-1},c) \parallel p_{\theta, c}(\bm{z}_L)))$.
    \label{eq:kl}
}
\end{equation}

\miniheadline{Contrastive Loss}
\label{mini:cl}
The contrastive loss consists of two terms, positive pair loss $\mathcal{L_+}$, which encourages proximity between samples belonging to the same class, and negative pair loss $\mathcal{L_-}$, that penalizes proximity between samples of different classes, ensuring inter-class separation.
We define boolean matrices $P$ and $N$ for positive pairs and negative pairs, respectively, as
$
    P_{ij} =
    \begin{cases} 
    1 & \text{if } l_i = l_j \text{ and } i \neq j, \\
    0 & \text{otherwise}
    \end{cases}
$
and
$
    N_{ij} =
    \begin{cases} 
    1 & \text{if } l_i \neq l_j, \\
    0 & \text{otherwise},
    \end{cases}
$
with $l_i$ and $l_j$ being the labels of patches $i$ and $j$, respectively.
These loss terms then become
$
    \mathcal{L_+} = \frac{1}{\sum_{i,j}P_{ij}}\sum_{i,j}P_{ij}\cdot \mathcal{D}(\bm\mu^{(i)},\bm\mu^{(j)})
$
and
$
    \mathcal{L}_-=\sum_{i,j}N{ij}\cdot\ell_-(\mathcal{D}(\bm\mu^{(i)},\bm\mu^{(j)})),
$
with $\bm\mu^{(i)}$ being the predicted means of the posterior distribution over all hierarchy levels for a patch $i$ in batch $\bm{X}$, and $\mathcal{D}(\bm\mu^{(i)},\bm\mu^{(j)})$ a distance function.
In our experiments, we used the Euclidean distance.
Note that for $\mathcal{L_-}$ we define the penalty function 
$
    \ell_{-}(d) =
    \begin{cases} 
    0 & \text{if } d \geq m, \\
    (m - d)^2 & \text{otherwise}
    \end{cases},
$
with $m$ being the so-called \textit{margin}, a hyperparameter that must be set appropriately, \eg using grid-search.

The full contrastive loss term is finally defined as
\begin{equation}
    \label{eq:cl}
    \mathcal{L}_{CL} = \lambda\mathcal{L}_+ + (1-\lambda)\mathcal{L}_-,
\end{equation}
with $\lambda$ being a hyperparameter that balances the positive and negative pair loss with each other.

Readers might wonder why a contrastive loss is useful when a GMM prior is used, where for each structure to be classified (\ie for each label) we have defined a Gaussian component in its own right.
The main reason is that the GMM prior only takes effect at the uppermost hierarchy level $L$.
At all levels $i<L$, $\mathcal{L}_{CL}$ is taking care of the desired label-wise segregation of latent encodings.

\miniheadline{The Overall Loss of \epsSeg}
\label{mini:loss}
Taken all together, the overall loss of \epsSeg is
\begin{equation}
    \mathcal{L} = \mathcal{L}_{I}
        + \alpha_1 \mathcal{L}_{CE} 
        + \alpha_2 \mathcal{L}_{KL} 
        + \alpha_3 \mathcal{L}_{CL},
\end{equation}
where $\alpha_i$' are hyperparameters to adjust the contribution of each loss to one another.
We tuned those hyperparameters using grid search and manual tuning.


\noindent Next, we present empirical results obtained using \epsSeg and comparisons to several baseline methods on two dense EM datasets and one fluorescence microscopy dataset.

%% file: sec/4_results.tex
\section{Experiments and Results}
\label{sec:res}
\tabBaseline
\figSegModels

\miniheadline{Datasets}
We used the ``BetaSeg''~\cite{Muller2021-mg} dataset\footnote{\href{http://betaseg.github.io}{http://betaseg.github.io}}, which was made publicly available by the authors.
This Focused Ion Beam Scanning Electron Microscopy (FIB-SEM) dataset captured primary mouse pancreatic islet $\beta$ cells at a 16 nm isotropic resolution. 
The final dataset consists of two groups of high and low glucose cells and also provides human curated binary segmentation masks for seven subcellular structures, \ie centrioles, nucleus, plasma membrane, microtubules, golgi body, granules, and mitochondria.
Consistent with~\cite{maester}, we also chose the 4 high glucose cells for this work.
For evaluation, cells 1, 2, and 3 from four cell volumes of high glucose were used for training, while cell 4 served as an independent test set.

Next, we used the ``liver FIBSEM'' dataset, which consists of samples that were fresh needle biopsies fixed with 4\%PFA and 2\%GA in phosphate buffer. High contrast staining was performed with reduced osmium and Waltons lead aspartate stain~\cite{walton} and embedded in Epon. 
Sample preparation and imaging were done on a ZEISS GeminiSEM according to prior reports~\cite{xu}.
The final dataset consists of one cell volume with $11$ crops that have been extracted from a cell volume, annotated manually, and used for training, validation, and testing. The segmentation masks consist of six subcellular structures, mitochondria, peroxisomes, lipofuscin, basolateral membrane, open bile canaliculi, and closed bile canaliculi, along with an additional ``background'' category. 

\tabComboFluLabkit

\vspace{1em}
\tabZerial

\vspace{1em}
\tabComboLabelEntropy

\figSegModelsZerial
\figFlu

\vspace{5em}

While it is true that FIB-SEM datasets like ``BetaSeg''~\cite{Muller2021-mg} offer isotropic resolution suitable for 3D processing, this is not always the case in EM imaging, where data often come in 2D slices (especially in higher-throughput screens).

Furthermore, we conducted an experiment on the overlapping subset of two datasets Aitslab-bioimaging1~\cite{fludata1} and Aitslab-bioimaging2~\cite{fludata2}.
The Aitslab-bioimaging1 dataset is a benchmarking fluorescence microscopy dataset containing 50 images of Hoechst 33342-stained U2OS osteosarcoma cell nuclei, including annotations for nuclei, nuclear fragments, and micronuclei, designed for training and evaluating neural networks for instance and semantic segmentation and the Aitslab-bioimaging2 dataset is a fluorescence microscopy dataset containing 60 images of EGFP-Galectin-3 labeled U2OS osteosarcoma cells with hand-annotated cell outlines, designed for training and benchmarking neural networks for instance and semantic segmentation, with over 2200 annotated cell objects and compatibility with object detection tasks.
The overlapping subset of them contains 30, 2-channel images for training and 10 for testing.

\miniheadline{Evaluation Metrics}
We used the Dice Similarity Coefficient (DSC) to evaluate the segmentation performance.
DSC is a widely used metric in image segmentation and measures the similarity between the predicted and actual segmentation masks.

Let $A$ and $B$ be two sets representing the binary segmentation masks of the ground truth and the predicted segmentation.
The Dice coefficient is defined as $Dice(A, B) = \frac{2|A\cap B|}{|A|+|B|}$, where $|A\cap B|$, the number of overlapping pixels between the predicted and ground truth masks, $|A|$, the number of pixels in the ground truth mask, and $|B|$, the number of pixels in the predicted mask.

\miniheadline{Experiments}
We use an architecture similar to the one used in the HDN work~\citep{hdn}.
For all hyperparameters we have introduced, we used grid searches to find a good balance between performance and stability.
We first evaluate our method on the ``BetaSeg'' dataset~\cite{Muller2021-mg} and compare its performance against baseline methods, as shown in Table~\ref{tab:baseline}.
They demonstrate that our approach outperforms existing baselines in terms of DSC (F1-score).
For the Labkit baseline, we trained per cell and show the results in Table~\ref{tab:labkit} and report the best class-wise performance in Table~\ref{tab:baseline}.
Quantitative segmentation results are shown in Figure~\ref{fig:seg_models} (complete Figure~\ref{fig:seg_models_full}).

To further validate the robustness of our method, we conduct experiments on the ``liver FIBSEM'' dataset, comparing it with U-Net baselines (fully and sparsely-supervised).
Quantitative and qualitative results are shown in Table~\ref{tab:zerial} and Figure~\ref{fig:seg_models_zerial}, respectively (complete Figure~\ref{fig:seg_models_zerial_full}).
Additionally, we show \epsSeg results on a fluorescent microscopy dataset (see Table~\ref{tab:flu} and Figure~\ref{fig:seg_models_flu}).

\miniheadline{Model Ablations}
We strip our model down to a vanilla HVAE and then re-introduce one component at a time, showing how each of the modules we have introduced above contributes to the overall performance we report.
These results on the ``BetaSeg'' dataset are shown in Table~\ref{tab:loss}.

Additionally, we evaluate how the quality of the results depends on the amount of available training labels.
To this end, we are starting from $0.05$\% of the total image data available in the ``BetaSeg'' dataset and gradually decreasing the number of used training labels down to $0.0025$\%.
The results of these experiments can be found in Table~\ref{tab:label}.
As discussed in Section~\ref{sec:method}, $\mathcal{L}_H$ helps us gain additional performance also from the unlabeled data, which we measure and report in Table~\ref{tab:entropy}.
Finally, we measured the effect of differently sized masking regions in Table~\ref{tab:mask}.

\miniheadline{Limitations}
\label{minih:limit}
While \epsSeg achieves competitive segmentation results using only sparse supervision, several limitations remain.
First, all experiments we present are conducted on 2D images. 
Extending the presented framework to operate in full 3D is an important next step, especially for volume EM data analysis.
Second, we noticed that the effectiveness of our entropy-based loss must be improved, \eg by replacing it with a more adaptive or data-driven strategy.
Finally, in the presented form, hyperparameters such as the contrastive loss margin still require manual tuning, which is not ideal for ease of use by biological experts.

\vspace{1em}
\tabComboLossMasking

%% file: sec/5_discussion.tex
\section{Discussion}

Here we presented \epsSeg, a novel semantic segmentation approach that leverages the variational latent representation of hierarchical variational autoencoders (HVAEs) trained on a limited amount of pixel-labels in an inpainting setup.
We used a GMM prior instead of the traditionally employed Gaussian prior and introduced a novel segmentation head that incorporates both a cross-entropy loss and an entropy loss to leverage available data for which no ground truth (GT) class-labels are available.
The integration of contrastive loss, combined with the structural advantages of the GMM prior, provides a means to effectively distinguish biological structures directly from the latent space encoding.

Transformer-based architectures, as used in MAESTER~\cite{maester}, usually have a rather large number of trainable parameters (\ie $328,452,352$ trainable parameters in MAESTER).
This makes such approaches less applicable to life-scientists since they require rather powerful computing setups.
Even our biggest network, in contrast, only employs $3,800,869$ trainable parameters (see Tables~\ref{tab:resblock} and~\ref{tab:latent}), making it fast to train and easy to use.
Our experiments also highlight an interesting fact, namely that smaller mask sizes with consistent labels emerged as the best strategy. 
This stands in contrast to Transformer-based approaches, where a relatively large fraction of the input images is masked during training~\cite{maester}. 

By combining hierarchical representations with advanced regularization techniques such as contrastive learning, we have shown that we can achieve competitive segmentation performance on complex microscopy data, even with relatively small models and limited training data.
The proposed approach tackles the challenge of label scarcity, enhances latent space representations tailored to structured biological data, and lays the groundwork for future exploration of semi-supervised learning techniques and adaptive latent priors.

Overall, this work bridges the gap between fully supervised and unsupervised methods by offering a scalable approach for large-scale  biomedical semantic image data segmentation.

%% file: supplement.tex
\newpage

\begin{center}
    {\LARGE \textbf{\epsSeg: Sparsely Supervised Semantic Segmentation\\of Microscopy Data}}
\end{center}

\vspace{0.5em}

\begin{center}
    {\Large \textbf{\textit{Supplementary Material}}}
\end{center}

\setcounter{figure}{0} 
\renewcommand{\thefigure}{S\arabic{figure}} 

\setcounter{table}{0} 
\renewcommand{\thetable}{S\arabic{table}} 

\appendix

\figBase

\figCl

\FloatBarrier

\newpage

\miniheadline{Entropy-based Loss}
When the sample $\bm{y'}$ of the Gumbel-Softmax distribution is uniform, the network is maximally unsure about which class to predict for the current input patch.
We noticed that this is commonly the case early during training, where the network has not yet seen a lot of patches for which ground truth labels are available.

To encourage the network not to predict a uniform $\bm{y'}$, we introduced an entropy loss for all patches $\bm{x}^{(j)} \in \bm{X}$ for which we do not have a ground truth class label.

\begin{equation}
    \mathcal{L}_H = -\sum_{\bm{x}^{(j)}\in \bm{X}} \bm{y}'^{(j)}log(\bm{y}'^{(j)}).
\end{equation}

\tabBaselineSupplement

\tabResBlock

\tabLatent

\figSegModelsFull

\figSegModelsZerialFull

\tabLabelReb